\documentclass[journal]{IEEEtran}
\IEEEoverridecommandlockouts
\usepackage{cite, amsmath,amssymb,amsfonts, algorithmic, graphicx, subfig, epstopdf, textcomp, xcolor}
\usepackage{bbold, dsfont, multirow, multicol, bm, lineno}
\usepackage[ruled]{algorithm2e}
\def\BibTeX{{\rm B\kern-.05em{\sc i\kern-.025em b}\kern-.08em
    T\kern-.1667em\lower.7ex\hbox{E}\kern-.125emX}}
\usepackage{longtable}
\begin{document}

\title{\Huge Classification-Aided Robust Multiple Target Tracking Using Neural Enhanced Message Passing}

\author{{Xianglong Bai}\thanks{The authors are with the School of Automation, Northwestern Polytechnical University, Xi'an 710129, China, and the Key Laboratory of Information Fusion Technology, Ministry of Education, Xi'an, Shaanxi, 710072, China.
Zengfu Wang is also with the Research \& Development Institute of Northwestern Polytechnical University in Shenzhen, Shenzhen 518057, China.
This work was supported in part by the National Natural Science Foundation of China under Grant U21B2008, Grant 62233014, Grant 61873211, and in part by the Natural Science Basic Research Plan in Shaanxi Province of China under Grant 2021JM-06.}, Zengfu Wang*\thanks{*~Corresponding author:~Zengfu Wang.}, Quan Pan, Tao Yun, Hua Lan}

\maketitle

\begin{abstract}
We address the challenge of tracking an unknown number of targets in strong clutter environments using measurements from a radar sensor.
Leveraging the range-Doppler spectra information, we identify the measurement classes, which serve as additional information to enhance clutter rejection and data association, thus bolstering the robustness of target tracking.
We first introduce a novel neural enhanced message passing approach, where the beliefs obtained by the unified message passing are fed into the neural network as additional information.
The output beliefs are then utilized to refine the original beliefs.
Then, we propose a classification-aided robust multiple target tracking algorithm, employing the neural enhanced message passing technique. 
This algorithm is comprised of three modules: a message-passing module, a neural network module, and a Dempster-Shafer module.
The message-passing module is used to represent the statistical model by the factor graph and infers target kinematic states, visibility states, and data associations based on the spatial measurement information.
The neural network module is employed to extract features from range-Doppler spectra and derive beliefs on whether a measurement is target-generated or clutter-generated. 
The Dempster-Shafer module is used to fuse the beliefs obtained from both the factor graph and the neural network.
As a result, our proposed algorithm adopts a model-and-data-driven framework, effectively enhancing clutter suppression and data association, leading to significant improvements in multiple target tracking performance. 
We validate the effectiveness of our approach using both simulated and real data scenarios, demonstrating its capability to handle challenging tracking scenarios in practical radar applications.

\end{abstract}


\section{Introduction}\label{sec:INTRODUCTION}
\subsection{Background, Motivation, State of the Art}\label{subsec:Background, Motivation, State of the Art}
Multiple target tracking (MTT) using radars involves estimating the kinematic states of targets over a specific period, playing a significant role in both military and civilian fields.
The detect-then-track framework is commonly utilized for MTT, which first generates a list of measurements by a detector with a given threshold and then estimates the states of targets using a tracker.
Traditional MTT methods within this framework include the global neighbor association algorithm~\cite{2011Tracking}, the multi-hypothesis tracker~\cite{Blackman2004}, and the joint probabilistic data association filter~\cite{Bar1995Multitarget}.
More recently, methods based on finite set statistics, such as the cardinalized probability hypothesis density (CPHD) filter~\cite{Vo2007} and the multi-Bernoulli (MB) filter~\cite{Vo2009}, have been developed.
In environments with strong sea or ground clutter, using a higher threshold can cause target loss, while a lower threshold may result in an excessive number of false tracks.
Another approach, track-before-detect (TBD)~\cite{Grossi2013, Aprile2016}, utilizes multi-frame information to enhance the performance of small target detection in low signal-to-clutter ratio (SCR) environments.
Unfortunately, TBD methods often rely on the accumulated signal energy of the trajectory as the metric for target detection. This becomes problematic in cluttered environments where strong reflection points on the ground and sea surfaces can create clutter energy higher than that of the target. As a consequence, the performance of target detection and tracking is degraded.

In this paper, we explore the distinctions between target-generated and clutter-generated measurements in raw radar echo signals.
One crucial aspect we focus on is the range-Doppler (RD) spectra, which can be derived using standard radar signal processing techniques such as matched filtering and coherent integration~\cite{Richards2010}.
Specifically, the target spectra and the clutter spectra differ in terms of spectral power distribution, spectral phase fluctuations, spatial texture of echo power and spatial texture of the spectra~\cite{Li2014,Shi2019,Shui2020,Gao2021}.
For example, the RD spectra of target echoes exhibit a distinct sharp peak due to the fact that the power of target returns concentrates on several Doppler bins, particularly at the higher radar resolution~\cite{Gao2021}.
On the other hand, clutter echoes produce an obtuse peak in their RD spectra due to the presence of textures with structural trends and the power distribution on the wide main clutter region in the Doppler domain~\cite{Shui2020}.
Leveraging these differences, the RD spectra can serve as valuable additional information to enhance clutter rejection and data association processes, thereby significantly improving the performance and robustness of MTT~\cite{Shalom2005}.

Traditional methods use RD spectra to create feature vectors, followed by training classifiers to distinguish between target-generated and clutter-generated measurements.
Li~\emph{et al.}~\cite{Li2014} used a Bayesian classifier to identify ground clutter and weather signals using spectral features.
Shui~\emph{et al.}~\cite{Shui2020} developed a feature-based detector using seven salient features of radar returns to enhance the detection capability of high-resolution maritime radars for sea-surface small targets.
However, feature engineering often falls short in accurately describing the details of the real data generation process.
In contrast, learning-based methods can effectively extract all relevant information from the raw sensor data.
Gao~\emph{et al.}~\cite{Gao2021} proposed a detection scheme that leverages signal structure information in RD spectra, learned through a convolutional neural network (CNN).
Wen~\emph{et al.}~\cite{Wen2022} proposed a two-step detection framework using deep CNN on sequential RD spectra.
The intra-frame detection was achieved by identifying differences in features, while inter-frame detection involved correlations between moving targets and sea clutter.
However, the integration of RD spectral data with spatial measurement information for classification-assisted MTT has not yet been considered.

In recent studies, MTT problems have been approached as inference problems leveraging the probabilistic graphical models, and solved by Bayesian inference algorithms including belief propagation (BP)~\cite{Yedidia2005}, Variational Bayesian (VB)~\cite{Zhang2019} and unified message passing (MP)~\cite{Riegler2013}.
Initially, the BP algorithm gained popularity for data association~\cite{Williams2014, Williams2018, Sun2016IF} due to its effectiveness on hard constraints.
Subsequently, both the BP algorithm and its particle-based implementation found applications in scalable multi-sensor MTT algorithms~\cite{meyer2017scalable, Meyer2018, Soldi2019}.
Moreover, the BP method has been successfully applied to the labeled MB filtering~\cite{Kropfreiter2019}, cooperative self-localization and MTT~\cite{Sharma2019}, extended target tracking~\cite{Meyer2020,meyer2021scalable}, and other tracking and estimation problems~\cite{Leitinger2019a, Li2022, Cormack2019, Gaglione2022}.
The VB-based MTT algorithms were proposed in~\cite{Turner2014, Lau2016, Lan2019}, leveraging the structured mean-field (MF) approximation~\cite{Zhang2019} and exploiting a tractable family of distributions to approximate factorised distributions of the target states estimation, track management and data association.
Additionally, the unified MP method, combining the virtues of BP and MF while circumventing their drawbacks, has been used for MTT~\cite{Lan-2020-107621, Lan2020}.

Recently, inference methods have focused on the combination of probabilistic graphical models with neural networks (NNs) to make use of the advantages of both model-driven inference and data-driven inference~\cite{Johnson2016, Kuck2020, Satorras2019, Satorras2020}.
One prominent method in this domain is the neural enhanced belief propagation (NEBP) introduced by Satorras \emph{et al.}
~\cite{Satorras2019, Satorras2020}.
In order to correct errors introduced by cycles and model mismatch when using probabilistic graphical models, they established a graph NN (GNN) matching the factor graph (FG), allowing GNN messages to complement corresponding BP messages.
Liang~\emph{et al.}~\cite{Liang2021} used NEBP for cooperative localization, which complemented BP with learned information provided by a GNN.
Moreover, Liang~\emph{et al.}~\cite{Liang2022Fusion, Liang2022} extended NEBP to MTT where probabilistic data association is enhanced by learned information provided by a GNN.
For training the GNN, they presented a loss function comprising false alarm rejection and data association.
Additionally, Gaglione and Soldi~\emph{et al.}~\cite{Soldi2020, Gaglione2020} proposed an MTT framework based on BP,
exploiting information from a CNN that classified the low-frequency active sonar detection.

\subsection{Contributions and Paper Organization}\label{subsec:Contribution and Paper Organization}
Inspired by NEBP~\cite{Satorras2020}, in this paper, we present a novel classification-aided MTT method employing the neural enhanced message passing (NEMP).
NEMP is a hybrid inference algorithm, which combines the strengths of FG and NN into an integrated FG-NN.
The NEMP algorithm starts by running the unified MP~\cite{Riegler2013}.
Then, the NN takes the beliefs from MP and additional feature information as input, and outputs a refined version of the original beliefs.
Finally, the beliefs from both BP and NN are fused using the Dempster-Shafer (DS) rule~\cite{Liu-Zhun-Ga2018}.
%
%
Building upon NEMP, we propose a classification-aided MTT framework.
In model-based part, we use an FG to represent the statistical model of the MTT problem, and then run the unified MP to obtain the beliefs of probabilistic data association based on the spatial measurements.
In the data-driven part, a CNN takes the RD spectra as input and outputs its features.
Moreover, a multi-layer perceptron (MLP) takes the features and the beliefs of probabilistic data association from FG part as input, and outputs the refined beliefs that indicating whether measurement is target-generated or clutter-generated.
Finally, the DS rule, which provides a theoretical framework to model and fuse uncertain information, is employed to fuse the beliefs from BP and NN.
The resulting combined beliefs are then used to calculate the final target kinematic states and visibility states.
The proposed algorithm effectively addresses the MTT problem by integrating spatial measurement information with classification information learned from the RD spectra.
As a result, it significantly improves MTT performance in terms of clutter rejection and data association.
The key contributions of this paper are summarized as follows.
\begin{itemize}
  \item We propose an NEMP method.
  In this method, we use unified MP combined with BP and MF approximation, instead of solely relying on BP.
  The unified MP excels in handling both hard constraints and conjugate-exponential models.
  Additionally, we adapt an NN for measurement classification.
  We also employ the DS rule to fuse classification beliefs, which proves to be more effective for multi-source belief fusion~\cite{Liu-Zhun-Ga2018}.
  \item For the first time, we propose a classification-aided MTT method that utilizes the measured spatial information with RD spectra information provided by a radar sensor.
  This approach employs the RD spectra to obtain classification beliefs regarding whether a measurement is target-generated or clutter-generated. These beliefs serve as additional information to aid clutter rejection and data association.
  By introducing classification-aided information, we can reduce the detection threshold without generating a large number of false tracks, particularly in strong clutter environments.
  \item Our proposed method, referred to as CA-MTT-NEMP, utilizes the NEMP method to solve the classification-aided MTT problem.
  CA-MTT-NEMP can be essentially decomposed into estimations of target kinematic state, target visibility state, and data association, which are carried out by the MF approximation, BP, and NEMP, respectively.
  CA-MTT-NEMP adopts a model-and-data-driven mechanism to combine beliefs from both the FG part and the NN part.
  We have demonstrated that this novel mechanism has better performance compared to using model-based inference or data-driven inference alone.
\end{itemize}

In comparison to other classification-aided MTT methods utilizing BP and NN~\cite{Liang2022,Soldi2020},
this paper differs in the type of data used for classification.
While~\cite{Liang2022} uses the shape information provided by a light detection and ranging sensor and~\cite{Soldi2020} uses the range-bearing map provided by a low-frequency active sonar,
we employ RD spectra data provided by a radar.
Furthermore, the proposed method employs the NEMP technique,
distinguishing it from~\cite{Liang2022}, which uses the NEBP method, and~\cite{Soldi2020},
which relies solely on BP and NN without any information interaction.

The rest of this paper is organized as follows.
The system model and problem formulation of MTT are described in Section~\ref{sec:system}.
Section~\ref{sec:NEBP for RMTT} derives the proposed CA-MTT-NEMP algorithm.
Section~\ref{sec:EXPERIMENTAL} evaluates the performance of the proposed algorithm.
At last, Section~\ref{sec:CONCLUSIONS} concludes this paper.

\section{System Model and Problem Formulation}\label{sec:system}
In this section, we first introduce the system model and then state the classification-aided MTT problem to be solved.

\subsection{Model of Target State}\label{subsec:Model of Target State}
At time $k$, let ${\bm{x}}_{i,k} \in \mathds{R}^n$ and $s_{i,k}$, $i \in \mathcal{I} \triangleq \{1, \ldots, N_T\}$ denote the kinematic state and visibility state of target $i$, respectively, where $N_T$ is the maximum number of targets.
The target visibility state $s_{i,k}  \in \{0,1\}$ is a binary random variable, i.e., if the target $i$ is present, $s_{i,k}=1$; otherwise, $s_{i,k}=0$.
Define the joint kinematic state and the joint visibility state of target $i$ at time $k$ as $\bm{X}_{k}=\{ \bm{x}_{i,k} \}_{i=1} ^{N_T}$ and $\bm{S}_k=\{s_{i,k} \}_{i=1} ^{N_T}$, respectively.
Additionally, define the joint kinematic state sequence and the joint visibility state sequence of target $i$ from time $1$ to $K$ as $\bm{X}_{1:K}=\{ \bm{X}_{k} \}_{k=1} ^{K}$ and $\bm{S}_{1:K}=\{ \bm{S}_{k} \}_{k=1} ^{K}$, respectively.
Under the assumption that the kinematic state and visibility state of each target evolve independently with a first-order Markov process, the PDF of $\bm{X}_{1:K}$ and $\bm{S}_{1:K}$ can be written as follows:
\begin{equation}\label{equ:target transition}
\begin{split}
p(\bm{X} _{1:K}) =\prod_{i = 1}^{N_T} p(\bm{x}_{i,1}) \prod_{k = 2}^{K} p(\bm{x}_{i,k}|\bm{x}_{i,k-1}), \\
\end{split}
\end{equation}
where $p(\bm{x}_{i,1})$ is the prior PDF at time $1$ and $p(\bm{x}_{i,k}|\bm{x}_{i,k-1})$ represents the transition PDF of the target's kinematic state, which can be determined by the dynamic model of each target~\cite{Lan-2020-107621, Lan2019, Lan2020}, and
\begin{equation}\label{equ:target visibility}
p(\bm{S} _{1:K})= \prod_{i = 1}^{N_T} p({s}_{i,1}) \prod_{k = 2}^{K} p({s}_{i,k}|s_{i,k-1}), \\
\end{equation}
where $p({s}_{i,1})$ is the prior PDF represented by a Bernoulli distribution, and the transition PDF $p({s}_{i,k}|s_{i,k-1})$ is represented by the transition matrix $\bm{T}$.
The target visibility probability is employed for track management.
If the visibility probability $p(s_{i,k}=1)$ exceeds a track confirmation threshold $\delta $, the track is declared as a target; otherwise, it is declared as a false track.

\subsection{Model of Radar Signal}\label{subsec:Model of Radar Signal}
We consider that a radar is operating in dwelling mode, providing a sufficient observation time to obtain the integration gain of a moving target.
The radar echoes are organized into a matrix with $M$ range bins and $P$ pulses.
In line with~\cite{Shi2019, Wen2022}, the fluctuated target returns can be described as
\begin{equation}\label{equ:target returns}
\begin{split}
  s_t(m,p) = & A_0(m) a(p) \exp \left(j \varphi+j \frac{4 \pi R(p)}{\lambda}\right), \\  
\end{split}
\end{equation}
where
$m \in \{1, \ldots, M\}$ and $p \in \{1, \ldots, P\}$ represent the indices of the range bin and the pulse, respectively;
$A_0(m)$ is the target signal amplitude at the $m$th range bin;
$a(p)$ is a highly correlated positive stochastic sequence representing the slow change of the target's Radar Cross Section~(RCS) over each pulse~\cite{Shi2019};
$\varphi$ follows a uniform distribution between $-\pi$ and $\pi$ representing the random initial phase;
$R(p)$ represents the target radial distance at the $p$th pulse;
$\lambda$ represents the radar wavelength.
The target signal amplitude at the $m$th range bin is defined as $A_0(m) = \sqrt{P_t \omega(m) }$, where $P_t$ \textcolor{blue}{is} the total power of target and $\omega(m)$ is the ratio of the target's radial length in the $m$th range bin to the target's total radial length.
Here, we define $P_c$ as the statistical power of pure clutter over $P$ pulses and $M$ range bins, and SCR as the ratio of total power of the target to the clutter power.
Thus, the total power of \textcolor{blue}{a} target is defined as $P_t=10 ^ {\rm SCR / 10} P_c$.

\subsection{Measurement Model}\label{subsec:Detection and Measurement Model}
We adopt the detect-then-track framework, where we initially apply matched filtering and coherent integration to the echoes of multiple pulses to enhance the SNR and obtain the RD spectra.
Subsequently, a detector with a false alarm rate $P_{\rm FA}$, a cluster and a plot-extractor are utilized to generate a list of candidate measurements~\cite{Richards2010}.
Let $\bm{z}_{j,k}$ and $\bm{M}_{j,k}$ denote the spatial information and RD spectra of measurement $j$, $j \in \{1,\ldots,N_{M,k}\}$, where $N_{M,k}$ is the number of measurements at time $k$.
The spatial measurement, denoted as a 2-dimensional vector $\bm{z}_{j,k}=[r_{j,k}, \ f_{j,k}]^{\rm T}$, include the range measurement $r_{j,k}$ and the Doppler frequency measurement $f_{j,k}$.
Since prolonged coherent integration may lead to range migration and Doppler frequency migration~\cite{Huang2019}, the RD spectra is defined as $\bm{M}_{j,k}=(\zeta_{j,k}^{m,p})_{m=1,\ldots,N_m,p=1,\ldots,N_p}$, which contains the signal amplitudes within the measurement centroid and its surrounding cells.
Here, $N_m$ and $N_p$ represent the maximum number of range bins and Doppler channels occupied by the target signal, respectively.
The spatial measurement set and RD spectra set at time $k$ are denoted as $\bm{Z}_{k}= \{\bm{z}_{j,k}\}_{j=1}^{N_{M,k}}$ and $\bm{M}_{k}= \{\bm{M}_{j,k}\}_{j=1}^{N_{M,k}}$, respectively.
The spatial measurement sequence and RD spectra sequence from time $1$ to time $K$ are represented as ${\bm{Z}} _{1:K}=\{\bm{Z}_{k}\}_{k=1} ^{K}$ and ${\bm{M}} _{1:K}=\{\bm{M}_{k}\}_{k=1} ^{K}$, respectively.

We introduce the joint data association events between measurements and targets as $\bm{A}_k=\{a_{i,j,k}\}_ {i=1} ^{N_T} \ {} _{j=0} ^{N_{M,k}}$.
For $i>0$, the binary association variable $a_{i,j,k}=1$ if measurement $\bm{z}_{j,k}$ is generated by target $i$; otherwise, $a_{i,j,k}=0$.
Notably, $a_{i,0,k}=1$ if target $i$ is not detected; otherwise, $a_{i,0,k}=0$, and $a_{0,j,k}=1$ if measurement $j$ is a false alarm; otherwise, $a_{0,j,k}=0$.
The joint data association sequence from time $1$ to time $K$ is denoted as $\bm{A} _{1:K} = \{\bm{A} _{k}\}_{k=1} ^{K}$.

The conditional distribution of the spatial measurement sequence ${\bm{Z}} _{1:K}$ given the target state sequence $\bm{X} _{1:K}$ and data association sequence $\bm{A} _{1:K}$ can be described as
\begin{equation}\label{equ:spatial likelihood function}
\begin{split}
  & p(\bm{Z}_{1:K}|\bm{X}_{1:K},\bm{A}_{1:K}) \\
= & \prod_{k=1}^{K} \prod_{j=1}^{N_{M,k}} P_{\rm FA}^ {a_{0,j,k}}
  \prod_{i=1}^{N_T} p(\bm{z}_{j,k}|\bm{x}_{i,k}) ^ {a_{i,j,k}}.
\end{split}
\end{equation}

The RD spectra-based classification can effectively distinguish targets and clutter~\cite{Shui2020, Li2013}.
The probabilistic discriminative model of RD spectra-based classification can be described by the conditional probability distribution of the data association sequence $\bm{A} _{1:K}$ given the RD spectra sequence ${\bm{M}} _{1:K}$, that is
\begin{equation}\label{equ:RD likelihood function}
p(\bm{A}_{1:K}|\bm{M}_{1:K})
=
\prod_{k=1}^{K} \prod_{j=1}^{N_{M,k}} p(a_{0,j,k} | \bm{M}_{j,k}).
\end{equation}
The probabilistic discriminative model obtains the conditional probability distribution $p(a_{0,j,k} | \bm{M}_{j,k})$ in an inference stage and subsequently uses this distribution to make optimal classification decisions.

\subsection{The Prior PDF of Data Association}\label{subsec: Data Association}
The joint prior probability of data association sequence $\bm{A}_{1:k}$ given the target visibility state sequence $\bm{S}_{1:k}$ is~\cite{Lan-2020-107621, Lan2019, Lan2020}
\begin{equation}\label{pirAonS}
\begin{split}
  p(\bm{A}_{1:K}& |\bm{S}_{1:K}) = \prod_{k=1}^{K} p(\bm{A}_{k}|\bm{S}_{k}) \\
 & = \prod_{k=1}^{K} \prod_{i=1}^{N_T} P_{\rm d} (s_{i,k}) ^{1 - a_{i,0,k}} (1 - P_{\rm d}(s_{i,k}))^{a_{i,0,k}},
\end{split}
\end{equation}
where $P_{\rm d} (s_{i,k}=1)=P_{\rm d}$ and $P_{\rm d} (s_{i,k}=0)=\varepsilon (0 < \varepsilon \ll 1)$ represent the detection probability of target $i$ given the visibility state $s_{i,k}$.

For a valid joint data association event $\bm{A}_k$, it must satisfy the following two constraints:
(a) Each measurement is originated from at most one target, that is, ${I_{j,k}}({\bm{\bar{A}} _{j,k}}) = 1$ if $ \sum_{{a_{i,j,k}} \in {\bm{\bar{A}} _{j,k}}} {{a_{i,j,k}}}  = 1 $; otherwise, ${I_{j,k}}({\bm{\bar{A}} _{j,k}}) = 0$, where ${\bm{\bar{A}} _{j,k}} = \{ a_{i,j,k}\} _{i = 0}^{N_T}$;
(b) Each target can generate at most one measurement, that is, $E_{i,k} (\bm{\tilde{A}} _{i,k}) = 1$ if $\sum_{a_{i,j,k} \in \bm{\tilde{A}} _{i,k}}{a_{i,j,k}} = 1$; otherwise $E_{i,k} (\bm{\tilde{A}} _{i,k}) = 0$, where $\bm{\tilde{A}} _{i,k} = \{ a_{i,j,k}\} _{j = 0}^{N_{M,k}}$.
Accordingly, define the following set of constraints.
\begin{equation}\label{equ:set of constraints1}
  \begin{aligned}
    & {I}({\bm{A} _{1:K}}) = \prod _{k=1}^{K} \prod _{j=1}^{N_{M,k}} {I_{j,k}}({\bm{\bar{A}} _{j,k}}), \\
  \end{aligned}
\end{equation}
\begin{equation}\label{equ:set of constraints2}
  \begin{aligned}
    {E}(\bm{A} _{1:K}) & = \prod _{k=1}^{K} \prod _{i=1}^{N_T} E_{i,k}(\bm{\tilde{A}} _{i,k}).
  \end{aligned}
\end{equation}

\subsection{Problem Statement}\label{subsec:Problem Statement}
The aim of classification-aided MTT is to simultaneously estimate $\bm{X}_{1:K}$ (target tracking) and $\bm{S}_{1:K}$ (target detection) given spatial measurements $\bm{Z}_{1:K}$ and RD spectra $\bm{M}_{1:K}$ with unknown data association $\bm{A}_{1:K}$.
This problem poses several challenges: 1) Optimal estimation of $\bm{X}_{1:K}$ and $\bm{S}_{1:K}$ has an exponential complexity due to the need to marginalize over the data association; 2) The conditional probability distribution $p(\bm{M}_{1:K}|\bm{A}_{1:K})$, which evaluates the probability of a measurement being target-generated or clutter-generated, is hard to model due to the influence of the complex environment on the RD spectra.
To address these challenges, we employ NEMP for classification-aided MTT in the next section.
The NEMP method provides a promising solution to efficiently estimate the target tracking and target detection while handling the uncertainties and complexities associated with the data association and RD spectra modeling.

\section{NEMP for Classification-Aid MTT}\label{sec:NEBP for RMTT}
In this section, we begin by introducing the NEMP method.
Then, we present the framework of the proposed NEMP method for classification-aided MTT.
Next, we delve into the details of the three main modules of the propose algorithm: the MP module, the NN module and the DS module.
We also give the complete NEMP method for data association.
Finally, we discuss the initialisation, implementation and the computational complexity of the proposed algorithm.

\subsection{NEMP}\label{subsec:NEMP}
NEMP implements a hybrid inference model comprising of NN and unified MP.
Let $\mathcal{G}_f = (\mathcal{V}_f, \mathcal{E}_f)$ be an FG, which contains two type of nodes $\mathcal{V}_f=\mathcal{V} \cup \mathcal{F}$, where $v_x \in \mathcal{V}$ denotes variable-nodes and $v_f \in \mathcal{F}$ denotes factor-nodes.
The graph also includes two types of edges: edges going from factor-nodes to variable-nodes and edges going from variable-nodes to factor-nodes.
The NEMP method can be divided into three main steps:
1) Run MP on the FG and pass the messages to NN; 2) Run NN; 3) Refine MP beliefs by fusing beliefs obtained from NN.
These three steps are repeated $T$ times.
After these iterations, the refined BP beliefs are used to calculate the marginal probability.
The NEMP algorithm can be summarized as follows:
\begin{equation}\label{equ:NEBP algorithm MTT}
  \begin{split}
    \bm{\tilde{\mu}}_{f \rightarrow x} ^t, \bm{\tilde{\mu}}_{x \rightarrow f} ^t & = {\rm{MP}} (\bm{{\mu}}_{f \rightarrow x} ^t), \\
    \bm{m}_{f \rightarrow x} ^t  & = {\rm{FG-NN}}(\bm{h}_x, \bm{\tilde{\mu}}_{f \rightarrow x} ^t), \\
    \bm{{\mu}}_{f \rightarrow x} ^{t+1} & = {\rm{DS}} (\bm{m}_{f \rightarrow x} ^t, \bm{\tilde{\mu}}_{f \rightarrow x} ^t).
  \end{split}
\end{equation}
In \eqref{equ:NEBP algorithm MTT}, $\rm{MP} (\cdot)$ represents the MP update equations~\cite{Riegler2013}, taking the factor-to-node messages $\bm{{\mu}}_{f \rightarrow x}^t$ as inputs and yielding computed results $\bm{\tilde{\mu}}_{f \rightarrow x} ^t$ and $\bm{\tilde{\mu}}_{x \rightarrow f} ^t$.
In \eqref{equ:NEBP algorithm MTT}, $\rm{FG-NN} (\cdot)$ runs the FG-NN equations.
It takes as input the node embedding $\bm{h}_{x}$, which comprises the variable-node embedding $\{ h_x|x \in \mathcal{X} \}$, and the messages $\bm{\tilde{\mu}}_{f \rightarrow x} ^t$ calculated by $\rm{MP} (\cdot)$.
It produces the latent beliefs $\bm{m}_{f \rightarrow x}^t$.
Finally, ${\rm{DS}}(\cdot)$ takes as input the beliefs $\bm{m}_{f \rightarrow x}^t$ and the messages $\bm{\tilde{\mu}}_{f \rightarrow x} ^t$ calculated by $\rm{MP} (\cdot)$, and provides a refinement for the current message estimates $\bm{\tilde{\mu}}_{f \rightarrow x}^{t+1}$.
After running the algorithm for $N$ iterations, the marginal distribution of $\hat{p}(x_i)$ can be obtained by multiplying all incoming messages to node $x_i$.

There are three main differences between NEMP used in this paper and NEBP.
Firstly, instead of BP, we employ unified MP combining BP and MF approximation~\cite{Riegler2013}, as referred to in the first equation of \eqref{equ:NEBP algorithm MTT}.
Secondly, instead of a GNN, the NN part consists of a CNN used for feature extraction network and an MLP used for classification, as referred to in the second equation of \eqref{equ:NEBP algorithm MTT}.
Thirdly, instead of using a simple summation of beliefs, we use DS rules to fuse the belief of BP and FG-NN, as referred to in the third equation of \eqref{equ:NEBP algorithm MTT}.

\subsection{Framework of the Proposed Method}\label{subsec:Framework of the Proposed Method}
The proposed NEMP method for classification-aided MTT is presented as a flow diagram in Fig.~\ref{fig:Factor graph}, providing an overview of the approach.
NEMP comprises three key modules, as follows:
\subsubsection{MP Module} The statistical model of the MTT problem is described by a joint PDF, which can be factorized and represented using an FG, as shown in the FG part of Fig.~\ref{fig:Factor graph}.
Then, we run MP on the FG, obtaining the messages of prediction, measurement evaluation and probabilistic data association.
After receiving the final beliefs of data association from the DS module, the estimations of target kinematic state and visibility state are performed in the MP module.
The details will be given in Section~\ref{subsec:BP Module}.
\subsubsection{NN Module}
\begin{figure}[!htbp]
  \centering
  \includegraphics[width=8cm]{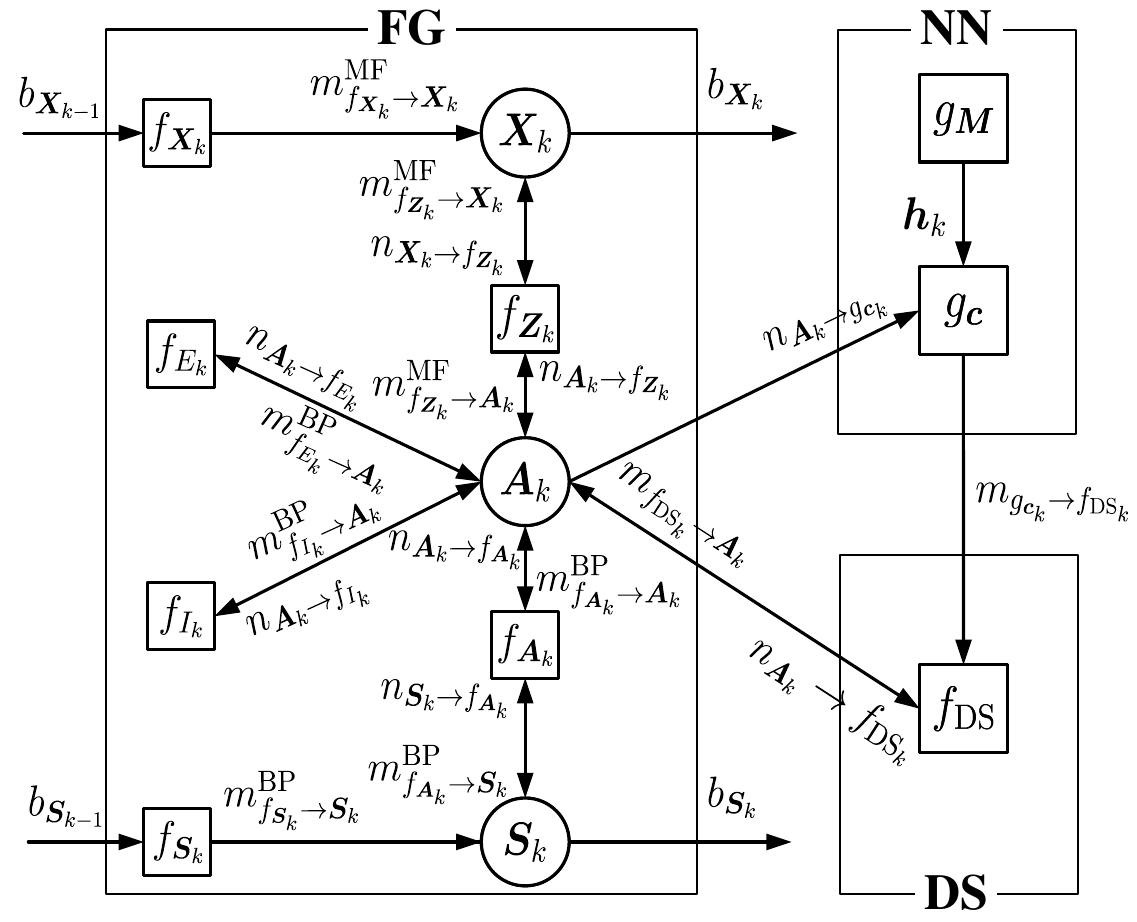}\\
  \caption{Flow diagram of the classification-aided MTT method.}\label{fig:Factor graph}
\end{figure}

The NN module primarily serves for RD spectral feature extraction and classification, as shown in the NN part of Fig.~\ref{fig:Factor graph}.
We construct a CNN used for feature extraction, which takes the RD spectra as input and generates low-dimensional features as output.
Following that, we design an MLP used for measurement classification, where the input is the extracted features and the beliefs of probabilistic data association from the MP module, and the output is the conditional probability indicating whether the measurement is target-generated or clutter-generated.
The proposed NNs are trained in a supervised manner, and to achieve this, a training set is employed, comprising RD spectra along with their corresponding labels.
The details will be given in Section~\ref{subsec:NN Module}.
\subsubsection{DS Module}
The DS module is used to fuse the beliefs of probabilistic data association from the MP module and the beliefs generated by the NN module.
This DS rule provides a theoretical framework to effectively model and fuse uncertain information, allowing us to fuse the beliefs from both MP and NN.
The details will be given in Section~\ref{subsec:Combination Module}.

Next, we will delve into the detailed derivations of the three modules.

\subsection{MP Module}\label{subsec:BP Module}
Let $\bm{\Theta} = \{ \bm{X}_{1:K}, \bm{S}_{1:K} , \bm{A}_{1:K} \}$ denote the collection of all the latent variables.
The joint posterior PDF $\mathcal{L} (\bm{\Theta}| \bm{Z}_{1:K})$ can be factorized as
\begin{equation}\label{equ:The Joint PDF}
\begin{split}
  \mathcal{L} (\bm{\Theta}| \bm{Z}_{1:K}) = & \frac{p(\bm{\Theta}, \bm{Z}_{1:K})}{p(\bm{Z}_{1:K})} \\
\propto & p(\bm{Z}_{1:K}|\bm{X}_{1:K},\bm{A}_{1:K}) p(\bm{X}_{1:K}) p(\bm{S}_{1:K})\\
  & \times p(\bm{A}_{1:K}|\bm{S}_{1:K}) {I}({\bm{A} _{1:K}}) {E}({\bm{A} _{1:K}}).
\end{split}
\end{equation}
Insert~\eqref{equ:target transition} for $p(\bm{X}_{1:K})$, \eqref{equ:target visibility} for $p(\bm{S}_{1:K})$, \eqref{equ:spatial likelihood function} for $p(\bm{Z}_{1:K}|$ $\bm{X}_{1:K},\bm{A}_{1:K})$, \eqref{pirAonS} for $p(\bm{A}_{1:K}|\bm{S}_{1:K})$, \eqref{equ:set of constraints1} for ${I}({\bm{A} _{1:K}})$, and \eqref{equ:set of constraints2} for ${E}({\bm{A} _{1:K}})$, yielding the factorization of~\eqref{equ:The Joint PDF} as
\begin{equation}\label{equ:The Joint factorization PDF}
\begin{split}
  \mathcal{L} & (\bm{\Theta}| \bm{Z}_{1:K}) \propto \prod_{k=1}^{K} \prod_{j=1}^{N_{M,k}} P_{\rm FA}^ {a_{0,j,k}} \prod_{i=1}^{N_T} p(\bm{z}_{j,k}|\bm{x}_{i,k}) ^ {a_{i,j,k}} \\
  & \times \prod_{i = 1}^{N_T} p(\bm{x}_{i,1}) p({s}_{i,1}) \prod_{k = 2}^{K} p(\bm{x}_{i,k}|\bm{x}_{i,k-1}) p({s}_{i,k}|{s}_{i,k-1}) \\
  & \times \prod_{k=1}^{K} \prod_{i=1}^{N_T} P_{\rm d} (s_{i,k}) ^{1 - a_{i,0,k}} (1 - P_{\rm d}(s_{i,k}))^{a_{i,0,k}} \\
  & \times \prod _{k=1}^{K} \prod _{j=1}^{N_{M,k}} {I_{j,k}}({\bm{\bar{A}} _{j,k}}) \prod _{i=1}^{N_{T}} E_{i,k}(\bm{\tilde{A}} _{i,k}).
\end{split}
\end{equation}
By observing~\eqref{equ:The Joint factorization PDF}, we define the factor nodes
$f_{\bm{X}_k} = \prod_{i=1}^{N_{T}} f_{\bm{x}_{i,k}}$ with $f_{\bm{x}_{i,k}} = p(\bm{x}_{i,k}|\bm{x}_{i,k-1})$,
$f_{\bm{S}_k} = \prod_{i=1}^{N_{T}} f_{{s}_{i,k}}$ with $f_{{s}_{i,k}} = p({s}_{i,k}|{s}_{i,k-1})$,
$f_{\bm{Z}_k} = \prod_{j=1}^{N_{M,k}} \prod_{i=0}^{N_{T}} f_{\bm{z}_{j,k}^{i}}$ with $f_{\bm{z}_{j,k}^{i}} = P_{\rm FA}^ {a_{0,j,k}}$ for $i=0$ and $f_{\bm{z}_{j,k}^{i}} = p(\bm{z}_{j,k}|\bm{x}_{i,k}) ^ {a_{i,j,k}}$ for $i>0$,
$f_{\bm{A}_k} = \prod_{i=1}^{N_{T}} f_{\bm{A}_{i,k}}$ with $f_{\bm{A}_{i,k}} = P_{\rm d} (s_{i,k}) ^{1 - a_{i,0,k}} (1 - P_{\rm d}(s_{i,k}))^{a_{i,0,k}}$,
$f_{I_k} = \prod _{j=1}^{N_{M,k}} f_{I_{j,k}}$ with $f_{I_{j,k}} = {I_{j,k}}({\bm{\bar{A}} _{j,k}})$,
and $f_{E_k} = \prod _{i=1}^{N_T} f_{E_{i,k}}$ with $f_{E_{i,k}} = E_{i,k}(\bm{\tilde{A}} _{i,k})$,
the set of variable nodes $\mathcal{I} \triangleq \{ \bm{X}_k, \bm{S}_k, \bm{A}_k \}_{i=1} ^{K}$, and the set of factor nodes $\mathcal{F} \triangleq \{ f_{\bm{X}_k}, f_{\bm{S}_k}, f_{\bm{Z}_k}, f_{\bm{A}_k}, f_{I_k}, f_{E_k} \}_{i=1} ^{K}$.
The corresponding FG is illustrated in the FG part of Fig.~\ref{fig:Factor graph}.
The set of factor nodes $\mathcal{F}$ can be divided into a BP part and an MF part, that is, $\mathcal{F}_{\rm BP} = \{ f_{\bm{S}_k}, f_{\bm{A}_k}, f_{I_k}, f_{E_k} \}_{i=1} ^{K}$ and $\mathcal{F}_{\rm MF} = \{ f_{\bm{X}_k}, f_{\bm{Z}_k} \}_{i=1} ^{K}$.
Based on the splitting of $\mathcal{F}$, the sets of variable nodes in the BP part and MF part are $\mathcal{I}_{\rm BP} = \{ \bm{X}_k, \bm{S}_k, \bm{A}_k \}_{i=1} ^{K}$ and $\mathcal{I}_{\rm MF} = \{ \bm{X}_k, \bm{A}_k \}_{i=1} ^{K}$, respectively.
To obtain beliefs of latent variables, we propose to run the following message passing on the FG in Fig.~\ref{fig:Factor graph}.
By employing the standard MF-BP message passing rules, the beliefs $b(\bm{X}_{k})$ and $b(\bm{S}_{K})$ are obtained by performing the following five steps for each scan $k$:
\subsubsection{Prediction}  The prediction messages include target kinematic state prediction message $m_{f_{\bm{X}_k} \rightarrow \bm{X}_k} ^{\rm MF} = \prod _{i=1} ^{N_T} $ $m_{f_{\bm{x}_{i,k}} \rightarrow \bm{x}_{i,k}} ^{\rm MF}$ and target visibility state prediction message $m_{f_{\bm{S}_k} \rightarrow \bm{S}_k} ^{\rm BP}= \prod _{i=1} ^{N_T} m_{f_{{s}_{i,k}} \rightarrow {s}_{i,k}} ^{\rm BP}$, which can be calculated as
\begin{equation}\label{equ:F2VM}
\begin{split}
  & m^{\text{MF}}_{f_{{\bm{x}} _{i,k}} \rightarrow {\bm{x}} _{i,k}} \\
= & \exp \left( \int b({\bm{x}} _{i,k-1}) \ln p\left({\bm{x}} _{i,k} |{\bm{x}} _{i,k-1} \right) d{{\bm{x}} _{i,k-1} } \right),
\end{split}
\end{equation}
\begin{equation}\label{equ:A2e11T}
\begin{split}
  m^{\text{BP}}_{f_{s_{i,k}} \rightarrow s_{i,k}}
  = & \sum_{s_{i,k-1}} p ({s}_{i,k}|s_{i,k-1}) b({s}_{i,k-1}),
\end{split}
\end{equation}
where $b({\bm{x}} _{i,k-1})$ and $b({s}_{i,k-1})$ are calculated at time $k-1$.
The messages $m^{\text{MF}}_{f_{{\bm{x}} _{i,k}} \rightarrow {\bm{x}} _{i,k}}$ and $m^{\text{BP}}_{f_{s_{i,k}} \rightarrow s_{i,k}}$ can be derived according from the transition PDF of ${\bm{x}} _{i,k}$ and $s_{i,k}$, respectively.
\subsubsection{Measurement evaluation} The measurement evaluation messages include spatial measurement evaluation message $m_{f_{\bm{Z}_k} \rightarrow \bm{A}_k} ^{\rm MF} = \prod _{i=0}^{N_T} \prod _{j=1}^{N_{M,k}} m_{f_{\bm{z}_{j,k} ^{i}} \rightarrow {a}_{i, j,k}} ^{\rm MF} $ and target visibility evaluation message $m_{f_{\bm{A}_k} \rightarrow \bm{A}_k} ^{\rm BP} = \prod _{i=1} ^{N_T} m_{f_{\bm{A}_{i,k}} \rightarrow {a}_{i,0,k}} ^{\rm BP}$, which can be calculated as
\begin{equation}\label{equ:Yk2Ak0}
\begin{split}
    & m_{f_{\bm{z}_{j,k} ^{i}} \rightarrow {a}_{i, j,k}} ^{\rm MF} \\
  = &
  \begin{cases}
\exp \int n_{\bm{x}_{i,k} \rightarrow f_{\bm{z}_{j,k} ^{i}}} \ln p(\bm{z}_{j,k}|\bm{x}_{i,k}) ^{a_{i,j,k}} {d} \bm{x}_{i,k}, \!\!\! & i>0,
\\
P_{\rm FA}^ {a_{0,j,k}}, & i=0,
\end{cases}
\end{split}
\end{equation}
\begin{equation}\label{equ:Ak2AkA1}
  \begin{split}
    & m_{f_{\bm{A}_{i,k}} \rightarrow {a}_{i,0,k}} ^{\rm BP} = \sum_{{s}_{i,k}} p(a_{i,0,k}|s_{i,k}) n_{s_{i,k} \rightarrow f_{\bm{A}_{i,k}}}.
  \end{split}
\end{equation}
where $n_{\bm{x}_{i,k} \rightarrow f_{\bm{z}_{j,k} ^{i}}} = m^{\text{MF}}_{f_{{\bm{x}} _{i,k}} \rightarrow {\bm{x}} _{i,k}}$ is calculated in \eqref{equ:F2VM} and $n_{s_{i,k} \rightarrow f_{\bm{A}_{i,k}}} = m^{\text{BP}}_{f_{s_{i,k}} \rightarrow s_{i,k}}$ is calculated in \eqref{equ:A2e11T}.

\subsubsection{Probabilistic data association}
Here, we focus on probabilistic data association, which is part of NEMP data association in Section~\ref{subsec:NEMP for Data association}.
Note that NEMP is an iterative algorithm, but for the sake of simplicity, we omit the iteration index $t$ here.
The data association messages include $m_{f_{{E}_k} \rightarrow \bm{A}_k} ^{{\rm BP}} = \prod _{i=1}^{N_T} \prod _{j=0}^{N_{M,k}} m_{f_{{E}_{i,k}} \rightarrow a_{i,j,k}} ^{{\rm BP}}$ and $m_{f_{{I}_k} \rightarrow \bm{A}_k} ^{{\rm BP}} = \prod _{j=1}^{N_{M,k}} \prod _{i=0}^{N_T} m_{f_{{I}_{j,k}} \rightarrow a_{i,j,k}} ^{{\rm BP}}$, which can be calculated as
\begin{equation}\label{equ:A2AkE11}
\begin{split}
  &m ^{{\rm BP}}_{f_{{E}_{i,k}} \rightarrow a_{i,j,k}} = \!\!\!\!\!\! \sum_{{{{\bm{\tilde{A}}} _{i,k}}} \backslash \{ a_{i,j,k} \} } \!\!\!\!\!\!\!\!\! f_{{E}_{i,k} } ({{\bm{\tilde{A}}} _{i,k}}) \prod_{j'=1 \backslash j} ^{N_{M,k}} \!\! n_{{a_{i,j',k} \rightarrow f_{{E}_{i,k} }}},
\end{split}
\end{equation}
\begin{equation}\label{equ:A2AkI11}
\begin{split}
  m ^{{\rm BP}}_{f_{{I}_{j,k}} \rightarrow a_{i,j,k}} & = \!\!\!\!\!\!\sum_{{{{\bm{\bar{A}}} _{j,k}}} \backslash \{ a_{i,j,k} \} } \!\!\!\!\!\!\!\!\! f_{{I}_{j,k}} ({{{\bm{\bar{A}}} _{j,k}}}) \prod_{i'=1 \backslash i} ^{N_T} n_{{a_{i',j,k} \rightarrow f_{{I}_{j,k}}}},
\end{split}
\end{equation}
where $n_{{a_{i,j,k} \rightarrow f_{{E}_{i,k} }}} = m_{f_{\bm{z}_{j,k} ^{i}} \rightarrow {a}_{i, j,k}} ^{\rm MF} m_{f_{\bm{A}_{i,k}} \rightarrow {a}_{i,0,k}} ^{\rm BP} m ^{{\rm BP}}_{f_{{I}_{j,k}} \rightarrow a_{i,j,k}}$, $n_{{a_{i,j,k} \rightarrow f_{{I}_{j,k}}}} = m_{f_{\bm{z}_{j,k} ^{i}} \rightarrow {a}_{i, j,k}} ^{\rm MF} m_{f_{\bm{A}_{i,k}} \rightarrow {a}_{i,0,k}} ^{\rm BP} m ^{{\rm BP}}_{f_{{E}_{i,k}} \rightarrow a_{i,j,k}}$ if $i>0$; otherwise, $n_{{a_{0,j,k} \rightarrow f_{{I}_{j,k} }}} = m_{f_{{\rm{DS}}_k} \rightarrow {a}_{0, j,k}}$ and is initialised to $n_{{a_{0,j,k} \rightarrow f_{{I}_{j,k} }}} = m_{f_{\bm{z}_{j,k} ^{0}} \rightarrow {a}_{0, j,k}} ^{\rm MF}$ at the first iteration.
The messages $m ^{{\rm BP}}_{f_{{E}_{i,k}} \rightarrow a_{i,j,k}}$ and $m ^{{\rm BP}}_{f_{{I}_{j,k}} \rightarrow a_{i,j,k}}$ are initialized as one and updated via~\eqref{equ:A2AkE11} and~\eqref{equ:A2AkI11} iteratively.
The messages~\eqref{equ:A2AkE11} and~\eqref{equ:A2AkI11} can be simplified and the computational complexity is linear in the number of targets and the number of measurements. For details, the reader is refer to~\cite{Williams2014,Lan2019,meyer2017scalable}.
\subsubsection{Measurement update}
The measurement update message includes spatial measurement evaluation message $m_{f_{\bm{Z}_k} \rightarrow \bm{X}_k} ^{\rm MF} = \prod _{i=1}^{N_T} \prod _{j=1}^{N_{M,k}} m_{f_{\bm{z}_{j,k} ^{i}} \rightarrow \bm{x}_{i,k}} ^{\rm MF} $, and target visibility evaluation message $m_{f_{\bm{A}_k} \rightarrow \bm{S}_k} ^{\rm BP} = \prod _{i=1} ^{N_T} m_{f_{\bm{A}_{i,k}} \rightarrow {s}_{i,k}} ^{\rm BP}$, which can be calculated as
\begin{equation}\label{equ:F2VM2XT}
\begin{split}
  & m_{f_{\bm{z}_{j,k} ^{i}} \rightarrow \bm{x}_{i,k}} ^{\rm MF}
= \exp \sum_{{a}_{i,j,k}} n_{{a}_{i,j,k} \rightarrow f_{\bm{z}_{j,k} ^{i}}} \ln p(\bm{z}_{j,k}|\bm{x}_{i,k}) ^{a_{i,j,k}},
\end{split}
\end{equation}
\begin{equation}\label{equ:A2e12T1}
\begin{split}
  m_{f_{\bm{A}_{i,k}} \rightarrow {s}_{i,k}} ^{\rm BP} = \sum_{a_{i,0,k}} p(a_{i,0,k}| s_{i,k}) n_{a_{i,0,k} \rightarrow f_{\bm{A}_{i,k}}},
\end{split}
\end{equation}
where $n_{{a}_{i,j,k} \rightarrow f_{\bm{z}_{j,k} ^{i}}} = m_{f_{\bm{z}_{j,k} ^{i}} \rightarrow {a}_{i, j,k}} ^{\rm MF} m ^{{\rm BP}}_{f_{{I}_{j,k}} \rightarrow a_{i,j,k}} m ^{{\rm BP}}_{f_{{E}_{i,k}} \rightarrow a_{i,j,k}}$ for $j>0$ and $n_{a_{i,0,k} \rightarrow f_{\bm{A}_{i,k}}} = m_{f_{\bm{A}_{i,k}} \rightarrow {a}_{i,0,k}} ^{\rm BP} m ^{{\rm BP}}_{f_{{E}_{i,k}} \rightarrow a_{i,0,k}}$.
\subsubsection{Calculation of beliefs} Finally, we can calculate the beliefs of target kinematic state $b(\bm{X}_{k}) = \prod _{i=1} ^{N_T} b(\bm{x}_{i,k})$, and target visibility state $b(\bm{S}_{K}) = \prod _{i=1} ^{N_T} b({s}_{i,k})$, given by
\begin{equation}\label{equ:A2e12T2}
b(\bm{x}_{i,k}) = m^{\text{MF}}_{f_{{\bm{x}} _{i,k}} \rightarrow {\bm{x}} _{i,k}} m_{f_{\bm{z}_{j,k} ^{i}} \rightarrow \bm{x}_{i,k}} ^{\rm MF},
\end{equation}
\begin{equation}\label{equ:A2e12T3}
b({s}_{i,k}) = m^{\text{BP}}_{f_{s_{i,k}} \rightarrow s_{i,k}} m_{f_{\bm{A}_{i,k}} \rightarrow {s}_{i,k}} ^{\rm BP}.
\end{equation}
The message $b(\bm{x}_{i,k})$ can be recognized as a state space model, amenable to computation through the Kalman filter (KF)~\cite{Lan-2020-107621, Lan2019, Lan2020}.
On the other hand, the message $b({s}_{i,k})$ can be recognized as a hidden Markov model and can be effectively calculated using BP.

The derivations of the five steps of messages above are analogous to~\cite{Lan-2020-107621, Lan2019, Lan2020} and are omitted for the sake of brevity.

\subsection{NN Module}\label{subsec:NN Module}
In NN module, we utilize a CNN to extract features from RD spectra and we use an MLP to predict the conditional probability of whether a measurement is target-generated or clutter-generated.
The specific NN architectures employed in this process are described in detail below.

\subsubsection{The feature extraction NN}\label{subsubsec:The proposed CNN}
\begin{figure} [!htbp]
  \centering
  \includegraphics[width=7cm]{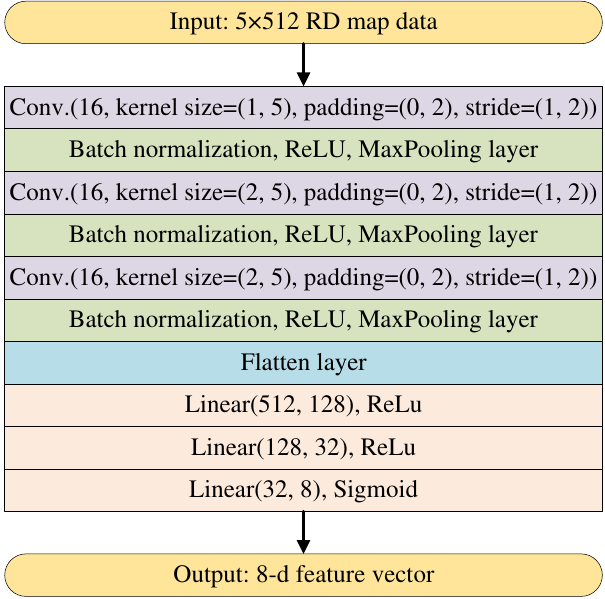} \\
  \caption{Architecture of the feature extraction network.}\label{fig:CNN_Architecture}
\end{figure}

For each measurement $j$, the RD spectra feature is extracted as $\bm{h}_{j,k} = g_{\bm{M}}(\bm{M}_{j,k})$, where $g_{\bm{M}}(\cdot)$ is a CNN.
The CNN is constructed by stacking multiple convolutional layers~\cite{Lecun2015}, allowing it to efficiently capture hierarchical representations from the RD spectra.
The architecture of the CNN designed is depicted in Fig.~\ref{fig:CNN_Architecture}, where the input to this network is an RD spectra with a size of $5 \times 512$, and the output is the feature vector of length $8$.
The CNN consists of three convolutional layers and three linear fully connected layers.
Each convolutional layer is followed by a batch normalization layer, a ReLU function and a MaxPooling layer to mitigate over-fitting, sensitivity and issues related to exploding and vanishing gradients.
Linear layer 1--2 is followed by the ReLU function.
Linear layer 3 is followed by the Sigmoid function, which transforms the features into the range $[0,1]$.
The set of RD spectra features at time $k$ is denoted as $\bm{h}_{k} = \{\bm{h}_{1, k}, \ldots, \bm{h}_{N_{M,k}, k}\}$.
\subsubsection{The classification NN}\label{subsubsec:The classification NN}
\begin{figure} [!htbp]
  \centering
  \includegraphics[width=7cm]{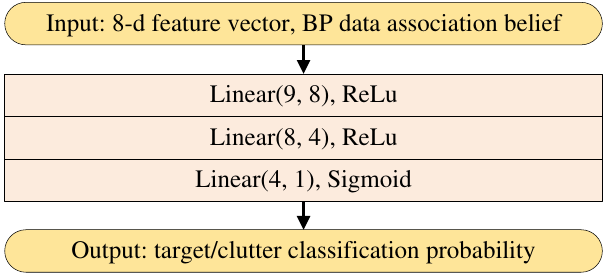} \\
  \caption{Architecture of the classification network.}\label{fig:MLP_Architecture}
\end{figure}

We obtain the classification belief for each measurement $j$ as follows,
\begin{equation}\label{equ:classification}
  m_{g_{c_{j,k}} \rightarrow f_{{\rm DS}_k}}  = g_c\left(\bm{h}_{j,k}, n_{a_{0, j,k} \rightarrow g_{c_k}}(1) \right),
\end{equation}
where $g_c(\cdot)$ represents an MLP show in Fig.~\ref{fig:MLP_Architecture}.
The MLP takes the input as the RD feature $\bm{h}_{j,k}$ obtained by the CNN and the beliefs of probabilistic data association $n_{a_{0, j,k} \rightarrow g_{c_k}}(1)$, and outputs the classification probability.
Here, we use probabilistic data association belief as additional information to aid in measurement classification.
Linear layer 1--2 is followed by the ReLU function, and Linear layer 3 is followed by the Sigmoid function, which transforms the classification probability into the range $[0,1]$.
The set of classification evidences at time $k$ is represented as $\bm{m}_{g_{c_k} \rightarrow f_{{\rm DS}_k}} = \{m_{g_{c_{1,k}} \rightarrow f_{{\rm DS}_k}}, \ldots, m_{g_{c_{N_{M,k},k}} \rightarrow f_{{\rm DS}_k}}\}$.

\subsubsection{Loss Function and Training}\label{subsubsec:Loss Function and Training}
The training process of the proposed NNs follows a supervised approach, assuming the availability of a training set comprising RD spectra and their corresponding labels.
The training is conducted in two steps.
The first step focuses on training the feature extraction network in Fig.~\ref{fig:CNN_Architecture}, where the output of the feature extraction network is fed into the second and third Linear layers of the classification network in Fig.~\ref{fig:MLP_Architecture}.
The second step focuses on training the classification network, where the output of the feature extraction network and the beliefs from FG part are fed into the classification network in Fig.~\ref{fig:MLP_Architecture}.
During training, the parameters of all NNs are updated using back-propagation, which computes the gradient of the loss function.
We use the following binary cross-entropy loss of measurements classification for both training steps,
\begin{equation}\label{equ:loss1}
  \mathcal{L} = \frac{-1}{N_{M,k}} \sum_{j=1}^{N_{M,k}} \omega_j^{\rm gt} \ln (\omega_j) +  (1 - \omega_j^{\rm gt}) \ln (1 - \omega_j),
\end{equation}
where $\omega_j = m_{g_{c_{j,k}} \rightarrow f_{{\rm DS}_k}}$ and $\omega_j^{\rm gt} \in \{0,1\}$ is the ground truth label for each measurement.
The ground truth label $\omega_j^{\rm gt}$ is assigned the value of 1 if the distance between the measurement and any ground truth position is smaller or equal to $T_{\rm dist}$, and 0 otherwise.

\subsection{DS Module}\label{subsec:Combination Module}
Finally, we use the DS rule to fuse the beliefs through two contributions: one that relies on the FG messages $n_{\bm{A}_k \rightarrow {f}_{{\rm DS}_k}} = \prod_{j=1}^{N_{M,k}} n_{a_{0, j, k} \rightarrow {f}_{{\rm DS}_k}}$, and one that is learned automatically during the process $m_{g_{{c}_k \rightarrow {f}_{{\rm DS}_k}}} = \prod_{j=1}^{N_{M,k}} m_{g_{{c}_{j,k} \rightarrow {f}_{{\rm DS}_k}}}$.
The combined belief $n_{{f}_{{\rm DS}_k} \rightarrow f_{\bm{A}_{k}}} = \prod_{j=1}^{N_{M,k}} n_{{f}_{{\rm DS}_k} \rightarrow f_{{a}_{0,j,k}}}$ is then returned to MP module, where it will be utilized in the data association step~\eqref{equ:A2AkE11} and~\eqref{equ:A2AkI11}.
The DS combination is achieved by following four steps.
Since the subsequent discussion applies to all measurements $j$ (where $j=1,\ldots,N_{M,k}$) and all time instances $k$ (where $k=1,\ldots,K$), the specific indices $j$ and $k$ are omitted for simplicity.

\subsubsection{Model of evidence reasoning}\label{subsubsec:Model of evidential reasoning}
In the context of evidence reasoning or belief function theory,
the RD spectra-based classification operates within a framework with a frame of discernment denoted as $\Omega = \{ \hbar, \ h\}$, which consists of two hypotheses.
Here, $\hbar$ represents the hypothesis that the measurement is clutter-generated, and $h$ represents the hypothesis that the measurement is target-generated.

\subsubsection{Construction of basic belief assignment}\label{subsubsec:Construction of basic belief assignment}
In the framework of evidence reasoning, the basic belief assignment~(BBA) is defined over the power set of $\Omega$, denoted by $2^{\Omega}$, which includes all subsets of $\Omega$.
The power set $2^{\Omega}$ contains $2^{|\Omega|}$ elements, including the empty set as well, written as $2^{\Omega} = \{ \emptyset, \{ \hbar\}, \{h\}, \Omega\}$.
A BBA is represented by a mass function $m(\cdot)$ from $2^{\Omega}$ to the interval $[0, 1]$, subject to the constrains that $m(\emptyset) = 0$ and $\sum_{A \in 2^{\Omega}} m(A) = 1$.
The subsets $A \in 2^{\Omega}$ for which $m(A) > 0$ are referred to as the focal elements of the BBA $m(\cdot)$.
In the context of pattern classification problems, the soft output of each classifier can be viewed as one source of belief represented by a BBA. The probabilistic output can be considered as the simple Bayesian BBA.
Given this definition of BBA, we consider two BBAs, $\bm{m}_1$ and $\bm{m}_2$, derived from the FG beliefs and NN beliefs, respectively.
Specifically, ${m}_1(\emptyset)=0$, ${m}_1(\hbar)=n_{a_{0} \rightarrow {f_{\rm DS}}}(0)$, ${m}_1(h)=n_{a_{0} \rightarrow {f_{\rm DS}}}(1)$, ${m}_1(\Omega)=0$, and ${m}_2(\emptyset)=0$, ${m}_2(\hbar)=1 - m_{g_{c} \rightarrow {f_{\rm DS}}}$, ${m}_2(h)=m_{g_{c} \rightarrow {f_{\rm DS}}}$, ${m}_2(\Omega)=0$.

\subsubsection{Combination by DS rule}\label{subsubsec:Combination by DS rule}
In evidence theory, the output of multiple classification results represented by BBA can be combined using the DS rule.
The combination of $\bm{m}_1$ and $\bm{m}_2$ by the DS rule is denoted as $\bm{m} = f_{\rm DS}(\bm{m}_1, \bm{m}_2) = \bm{m}_1 \oplus \bm{m}_2$ over $2^{\Omega}$, given by
\begin{equation}\label{equ:DS Rule}
\begin{split}
  m(A) =
\begin{cases}
\frac{\sum_{B \cap C=A} m_1(B) m_2(C)}{1-K}, & \forall A \in 2^{\Omega} \backslash \{ \emptyset \},
\\
0, & \text{if } A=\emptyset.
\end{cases}
\end{split}
\end{equation}
where $K=\sum_{B \cap C=\emptyset} m_1(B) m_2(C)$ is the total conjunctive conflicting masses, and $\oplus$ is the orthogonal sum.
The DS combination is an associative operation, meaning that the order in which the BBAs are combined does not affect the final combination result. This property ensures that the belief from different sources can be combined in any order without altering the overall result.

\subsubsection{Calculation of pignistic probability}\label{subsubsec:Calculation of pignistic probability}
In decision making, a BBA is typically transformed into a probability measure using the pignistic probability transformation denoted by $n_{{{f_{\rm DS}}} \rightarrow f_{{a}_{0}}}$.
The pignistic probability of the singleton class of $\omega$ is calculated as follows:
\begin{equation}\label{equ:pignistic}
  n_{{{f_{\rm DS}}} \rightarrow f_{{a}_{0}}}(\omega) = \sum_{X \in 2 ^{\Omega}, \omega \in X } \frac{1}{|X|} m(X),
\end{equation}
where $m(X)$ is the mass function associated with the BBA and $|X|$ represents the cardinality (number of elements) of set $X$.
These pignistic probabilities are used for data association in the FG part, completing the decision-making process based on the combined belief from both the FG beliefs and the NN beliefs.

\subsection{NEMP for Data Association}\label{subsec:NEMP for Data association}
Here, we present the complete NEMP method for data association.
As shown in Fig.~\ref{fig:Factor graph}, NEMP facilitates data association by means of iterative belief exchange between the MP, NN, and DS modules.
In the $t$th iteration, NEMP for data association encompasses the following steps:
\subsubsection{Measurement Classification} First, the classification belief for each measurement $j$ is obtained by~\eqref{equ:classification}.
The MLP takes the input as the RD feature $\bm{h}_{j,k}$ and the beliefs of probabilistic data association $n^{(t)}_{a_{0, j,k} \rightarrow g_{c_k}}(1)$, and outputs the classification probability $m_{g_{c_{j,k} \rightarrow f_{{\rm DS}_k}}}^{(t)}$.
\subsubsection{DS Combination} Next, we use the DS rule to fuse the beliefs from the FG messages $n^{(t)}_{a_{0, j,k} \rightarrow g_{c_k}}$ and classification beliefs $m_{g_{c_{j,k} \rightarrow f_{{\rm DS}_k}}} ^{(t)}$.
The fused belief $n_{{{f_{{\rm DS}_k}}} \rightarrow f_{{a}_{0,j,k}}} ^{(t)}$ can be calculated by~\eqref{equ:DS Rule}-\eqref{equ:pignistic}.
\subsubsection{Probabilistic data association} Finally, the fused belief $n_{{{f_{{\rm DS}_k}}} \rightarrow f_{{a}_{0,j,k}}} ^{(t)}$ is returned back to the MP module, where it will be utilized in the probabilistic data association by~\eqref{equ:A2AkE11} and~\eqref{equ:A2AkI11} iteratively.

This sequence of three steps is repeated for a total of $T$ iterations.
Consequently, the classification procedure embodies a model-and-data-driven framework, enhancing measurement classification performance.

\subsection{Initialization, Implementation, Computational Complexity}
The initial belief of target kinematic state $b ( \bm{X}_{1})$ is initialized by a two-point method~\cite{Oh2009}.
The belief of target visibility state is initialized as $b(s_{i,1}=1)=f_{\rm s}$ with $f_{\rm s}$ being the initial target visibility probability.

Given the initial beliefs $b(\bm{X}_{1})$ and $b(\bm{S}_{1})$, we can calculate $b(\bm{X}_{1:K})$ and $b(\bm{S}_{1:K})$ in principle by running CA-MTT-NEMP method described in Sec.~\ref{subsec:Framework of the Proposed Method}-Sec.~\ref{subsec:NEMP for Data association}.
The proposed CA-MTT-NEMP algorithm is summarized in Algorithm~\ref{algo:CA-MTT-NEMP}, which takes the beliefs of target kinematic state $b(\bm{X}_{k-1})$ and target visibility state $b(\bm{S}_{k-1})$ from time step $k-1$ as input and outputs the beliefs of target kinematic state $b(\bm{X}_{k})$ and target visibility state $b(\bm{S}_{k})$ at time step $k$.
The algorithm consists of the following steps:
Step 1) Calculate the prediction beliefs of target kinematic state $b(\bm{x}_{i, k|k-1})$ and target visibility state $b({s}_{i, k|k-1})$ as Line 1-3;
Step 2) Calculate the measurement evaluation for data association as Line 4-14;
Step 3) Extract RD spectra feature by CNN as Line 15;
Step 4) Perform data association by NEMP, which includes classification by MLP, beliefs combination, and BP data association iterations as Line 16-24;
Step 5) Calculate the measurement update messages as 25-30;
Step 6) Calculate the final beliefs of target kinematic state $b(\bm{x}_{i, k})$ and target visibility state $b({s}_{i, k})$ as Line 31-34.
\begin{algorithm}[!htbp]
{
\linespread{1.0} \selectfont
\caption{CA-MTT-NEMP Algorithm}\label{algo:CA-MTT-NEMP}
\LinesNumbered
\KwIn{$b(\bm{X}_{k-1}) = \{ \mathcal{N} (\bm{{x}}_{i, k-1}; \bm{\hat{x}}_{i, k-1},\bm{P} _{i, k-1}) \} _{i=1}^{N_T}$, $b(\bm{S}_{k-1}) = \{ b({s}_{i, k-1})\} _{i=1}^{N_T}$ from time $k-1$;}
\KwOut{$b(\bm{X}_{k}) = \{ \mathcal{N} (\bm{{x}}_{i, k}; \bm{\hat{x}}_{i, k}, \bm{P}_{i, k})\}$ , $b(\bm{S}_{k})=\{ b({s}_{i, k})\} _{i=1}^{N_T}$;}
\tcp{Prediction:}
\For{$i\leftarrow 1$ \KwTo $N_T$}
{
    $b(\bm{x}_{i, k|k-1}) = \mathcal{N} (\bm{{x}}_{i, k|k-1}; \bm{\hat{x}}_{i, k|k-1}, \bm{P} _{i, k|k-1})$ calculated by the prediction step of KF~\cite{Lan-2020-107621, Lan2019, Lan2020};
    $b({s}_{i, k|k-1}) = \bm{T} b({s}_{i, k-1})$;
}
\tcp{Measurement evaluation:}
\For{$j\leftarrow 0$ \KwTo $N_{M,k}$}
{
    \For{$i\leftarrow 0$ \KwTo $N_T$}
    {
        \uIf{$i = 0, j > 0$}
        {
            $ m_{f_{\bm{z}_{j,k} ^{0}} \rightarrow {a}_{0, j,k}} ^{\rm MF} = P_{\rm FA}^ {a_{0,j,k}}$;
        }
        \uElseIf{$i > 0, j = 0$}
        {
            $m_{f_{\bm{A}_{i,k}} \rightarrow {a}_{i,0,k}} ^{\rm BP} \!\!=\!\! \sum_{{s}_{i,k}} p(a_{i,0,k}|s_{i,k}) b({s}_{i, k|k-1})$;
        }
        \ElseIf{$i > 0, j > 0$}
        {
        $ m_{f_{\bm{z}_{j,k} ^{i}} \rightarrow {a}_{i, j,k}} ^{\rm MF} = \exp ( a_{i,j,k} \mathds{E}_{b(\bm{x}_{i, k|k-1})} [ \ln \mathcal{N} (\bm{z}_{j, k}; f(\bm{{x}}_{i, k}), \bm{R} _{k})] )$;
        }
    }
}
\tcp{RD feature extraction:}
\lFor{$j\leftarrow 1$ \KwTo $N_{M,k}$}
{
    Calculate $\bm{h}_{j,k} = g_{\bm{M}}(\bm{M}_{j,k})$
}

\tcp{NEMP for data association:}
\For{$t\leftarrow 1$ \KwTo $T$}
{
    \For{$j\leftarrow 1$ \KwTo $N_{M,k}$}
    {
        Calculate classification $m_{g_{c_{j,k}} \rightarrow f_{{\rm DS}_k}}$ as~\eqref{equ:classification}; \\
        Calculate DS combination $n_{{{f_{{\rm DS}_k}}} \rightarrow {{a}_{0,j,k}}}$ as \eqref{equ:pignistic};
    }
    \tcp{BP Data association:}
    \While{no convergence}{Update simplified messages of~\eqref{equ:A2AkE11} and \eqref{equ:A2AkI11} analogous to~\cite{Lan-2020-107621, Lan2019, Lan2020};}
}
\tcp{Measurement update:}
\For{$i\leftarrow 1$ \KwTo $N_{T}$}{
    $m_{f_{\bm{A}_{i,k}} \rightarrow {s}_{i,k}} ^{\rm BP} = \sum_{a_{i,0,k}} p(a_{i,0,k}| s_{i,k}) b(a_{i,0,k})$;

    \For{$j\leftarrow 1$ \KwTo $N_{M,k}$}{
        $m_{f_{\bm{z}_{j,k} ^{i}} \rightarrow \bm{x}_{i,k}} ^{\rm MF}=\mathcal{N}(\bm{z}_{j, k}; f(\bm{{x}}_{i, k}), \bm{R} _{k})^{\hat{a}_{i,j,k}}$;
    }
}
\tcp{Calculation of beliefs:}
\For{$i\leftarrow 1$ \KwTo $N_{T}$}{
    $b(\bm{x}_{i, k}) = \mathcal{N} (\bm{{x}}_{i, k}; \bm{\hat{x}}_{i, k}, \bm{P} _{i, k}) \propto b(\bm{x}_{i, k|k-1}) \prod_{i=1}^{N_{M,k}} \mathcal{N}(\bm{z}_{j, k}; f(\bm{{x}}_{i, k}), \bm{R} _{k})^{\hat{a}_{i,j,k}}$ calculated by the update step of KF~\cite{Lan-2020-107621, Lan2019, Lan2020}; \\
    $b({s}_{i,k})=\bm{T} b({s}_{i, k-1}) \sum_{a_{i,0,k}} p(a_{i,0,k}| s_{i,k}) b(a_{i,0,k})$.
}
}
\end{algorithm}

The computational complexity of the proposed CA-MTT-NEMP algorithm can be analyzed as follows.
The estimation of target kinematic state is carried out by the KF with a computational cost $c_{x} = \mathcal{O}(KN_T)$.
The estimation of target visibility state is carried out by the BP algorithm with a computational cost $c_{s}=\mathcal{O}(KN_T)$.
The data association is solved by LBP with a computational cost $c_{a}=\mathcal{O}(KN_aN_TN_M)$, where $N_a$ is the maximum number of BP iterations.
The RD spectra feature extraction and classification are carried out by a CNN and an MLP with computational costs $c_{\rm CNN}=\mathcal{O}(Ks_{\rm CNN}N_M)$ and $c_{\rm MLP}=\mathcal{O}(Ks_{\rm MLP}N_M)$, where $s_{\rm CNN}$ and $s_{\rm MLP}$ are constants that depend on the size and type of the CNN and MLP used.
Denote the total computational cost in NN part as $c_{m} = c_{\rm CNN} + c_{\rm MLP}$.
The belief combination is performed using the DS rule with a computational cost $c_{d}=\mathcal{O}(KN_M)$.
The overall computational complexity of the algorithm is given by $c_{\rm total} = c_{x}+c_{s}+c_a+c_{m}+c_{d}$.

\section{Simulation and Analysis}\label{sec:EXPERIMENTAL}
Next, we validate the performance of the proposed algorithm by introducing simulation targets into the real CSIR sea-clutter data.
We compare the proposed CA-MTT-NEMP algorithm against two other MP-based MTT algorithms: one algorithm performs MTT using measurements without classification aid, and the other algorithm performs MTT using measurements after clutter suppression by an RD spectra classifier.
For the sake of simplicity, we refer to the above three algorithms as NEMP, MP and MP-NN, respectively.
Additionally, we validate the generalization capability of the proposed method by testing it based on real IPIX radar sea-clutter data with simulated targets.

\subsection{Scenario Configuration}\label{subsec:Scenario Configuration}
\subsubsection{Dataset}
Our numerical evaluation is based on two primary datasets: the CSIR dataset collected during sea-clutter measurement trials at the Pretoria of South Africa~\cite{Wind2010}, and IPIX dataset collected by the McMaster University on the east coast of Canada in 1998~\cite{Bakker2019}.
The parameters for these datasets can be found in Table~\ref{tab:Radar Parameters of the IPIX and CSIR Data}.
To ensure a consistent pulse repetition frequency (PRF) across both sea-clutter datasets, we down-sampled the CSIR data with a step size of 5, adjusting the PRF from 5~kHz to 1 kHz.
For the IPIX radar data, we increase the number of range bins by employing repeat copies along the range direction.
Subsequently, we extract sets of experimental sea-clutter data, each containing M = 96 range bins and P = 10,000 pulses, from the adjusted IPIX and CSIR data.
In our simulation scenario, we consider a simulation duration of 150 seconds with a step size of 10 seconds, containing a total of 15 radar scans.
\begin{table} [!htbp]
\renewcommand \arraystretch{1.25}
  \centering
  \caption{Radar Parameters of IPIX and CSIR Data.} \label{tab:Radar Parameters of the IPIX and CSIR Data}
  \begin{tabular}{c|c|c}
  \hline
  \textbf{Parameters}        & \textbf{IPIX} & \textbf{CSIR} \\ \hline
  Center Frequency           & 9.39 GHz      & 9 GHz         \\ \hline
  Antenna Operation Mode     & stare         & stare         \\ \hline
  Pulse Repetition Frequency & 1 KHz         & 5 KHz         \\ \hline
  \end{tabular}
\end{table}

\subsubsection{Simulated targets}
\begin{figure} [!htbp]
  \centering
  \includegraphics[width=8cm]{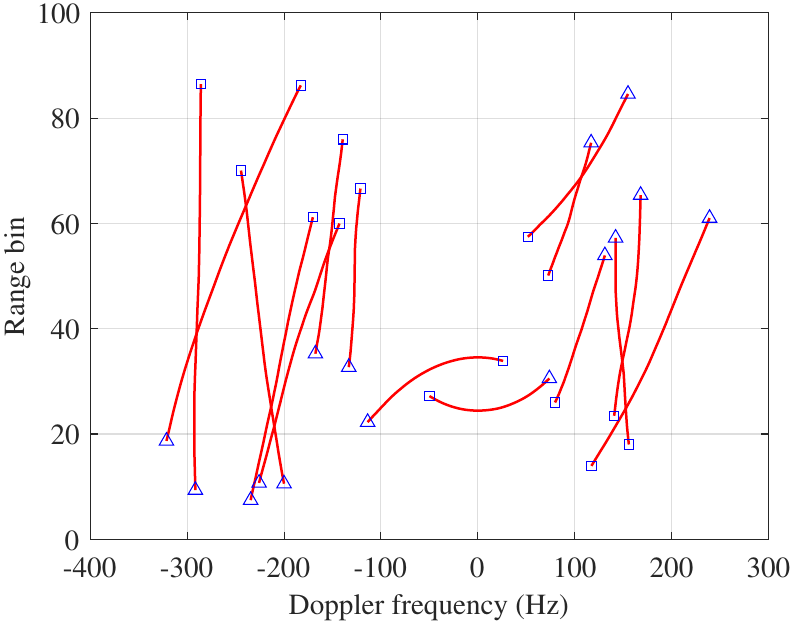} \\
  \caption{Some target trajectories in the RD domain. $\vartriangle$ and $\square$ represent the start point and the end point of a target, respectively.}\label{fig:True Trajectories of targets}
\end{figure}

\begin{figure}[!htbp]
  \centering
  \subfloat[Temporal domain\label{fig:Temporal domain1}]{\includegraphics[width=4.2cm]{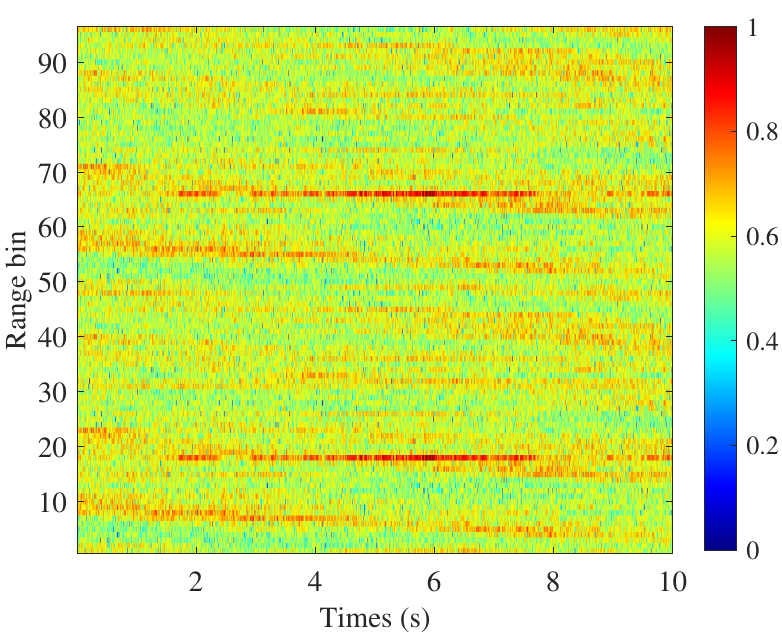}} \
  \subfloat[RD domain\label{fig:RD domain1}]{\includegraphics[width=4.2cm]{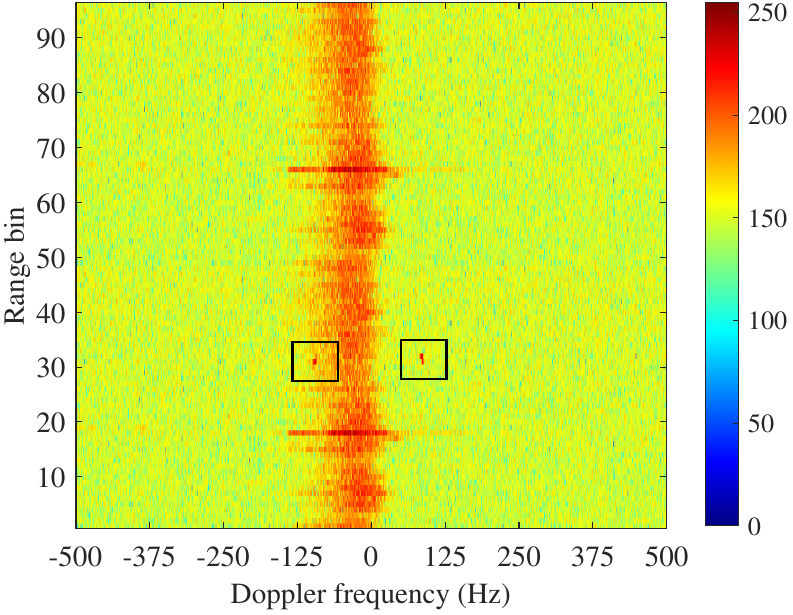}} \\
  \subfloat[Temporal domain\label{fig:Temporal domain2}]{\includegraphics[width=4.2cm]{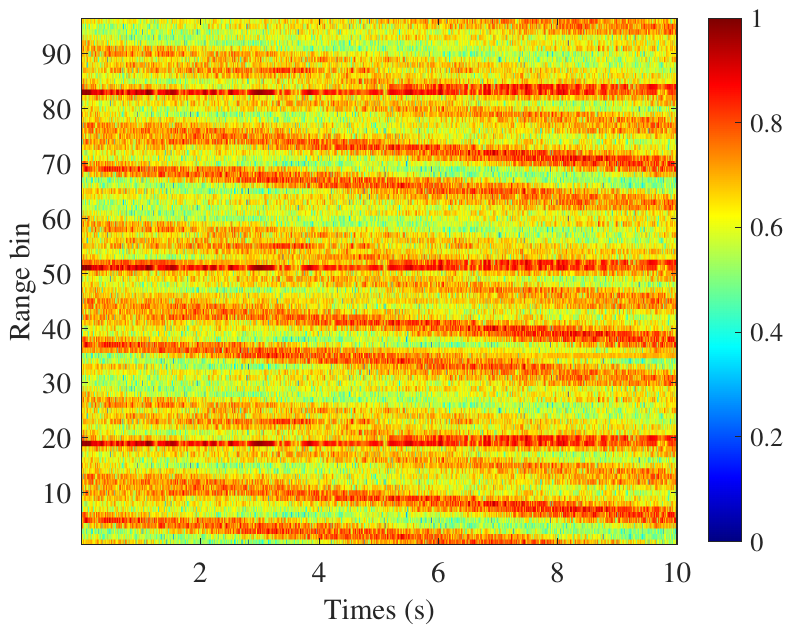}} \
  \subfloat[RD domain\label{fig:RD domain2}]{\includegraphics[width=4.2cm]{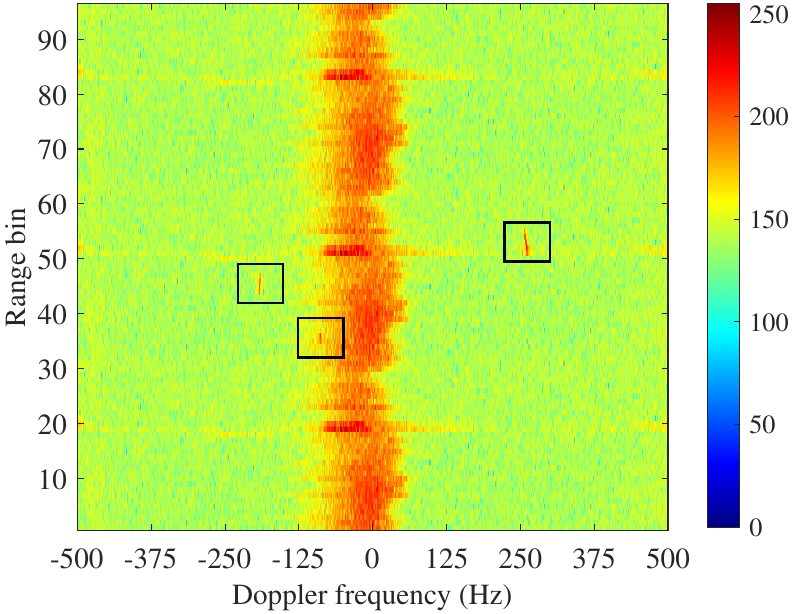}}
  \caption{Sea-clutter returns including simulated moving targets in the temporal and RD domains, respectively. In the RD domain, the targets are indicated by the black rectangle boxes.}\label{fig:Sea-clutter returns}
\end{figure}

The kinematic state of each target, denoted as $\bm{x}_{i,k}$, is characterized by its range, range velocity, and range acceleration.
The target follows a constant acceleration motion model, and the corresponding model parameters are
\begin{equation}\label{equ:cv}
  \bm{F} = \left[
             \begin{array}{ccc}
               1 & T & \frac{T^2}{2} \\
               0 & 1 & T \\
               0 & 0 & 1
             \end{array}
           \right]
, \ \
  \bm{Q} = \sigma_{a}^2 \left[
             \begin{array}{ccc}
               \frac{T^5}{20} & \frac{T^4}{8} & \frac{T^3}{6} \\
               \frac{T^4}{8}  & \frac{T^3}{3} & \frac{T^2}{2}\\
               \frac{T^3}{6}  & \frac{T^2}{2} & T\\
             \end{array}
           \right],
\end{equation}
where $T$ corresponds to the pulse repetition interval (PRI) with $\rm {PRI} = 1 / \rm {PRF}$, and $\sigma_{a} = 1 {\rm e}-4 \rm{m/s}^3$ represents the variance of the driving processes.
The total range length of the moving target is uniformly sampled between 5 m and 30 m.
We assume that all targets are present in 15 radar scans.
In accordance with~\cite{Wen2022}, the total radar returns are generated by adding simulated target returns into the cropped sea-clutter data with a specified SCR.
Subsequently, the echoes of multiple pluses are processed using fast Fourier transform (FFT) to obtain RD spectra.
Fig.~\ref{fig:True Trajectories of targets} displays some target trajectories in the RD domain, 
where it is seen that the Doppler frequency of the targets varies from -400 Hz to 300 Hz.
Meanwhile, there are also some targets that have a lower Doppler frequency and are located within the main sea clutter spectra.
Fig.~\ref{fig:Sea-clutter returns} provides two examples of sea-clutter returns containing simulated moving targets in the temporal and RD domains.
It is seen that the amplitudes of the simulated sea-surface small targets are fluctuated.

\subsubsection{Training and testing set}\label{subsubsec:Training and testing set}
In the training set, we randomly select sea-clutter backgrounds from 28 sets of real cropped CSIR data and introduce simulated moving targets for each radar scan.
FFT is first applied to the sea clutter returns to derive the RD spectra.
Subsequently, a constant false alarm rate (CFAR) detector is utilized to generate a list of candidate detection with a primary false alarm rate $P_{\rm FA}=0.28$.
To cluster candidate detection that are close together into the same detection region, we employ the DBSCAN algorithm~\cite{DBSCAN},
where the range bin clustering threshold is $R_{\rm th}$ and the Doppler bin clustering threshold is $D_{\rm th}$.
The spatial measurement is then calculated using the amplitude-weighted measurement centroid estimation algorithms, given by $ \bm{z}_{j,k} = {\sum_{\tau=1}^{N} A_{\tau} \bm{z}_{\tau,k}}/{\sum_{\tau=1}^{N} A_{\tau}}$, where $A_{\tau}$ is the amplitude measured in the $\tau$th primitive detection, and $N$ is the number of primitive detection of the detection region $j$.
To facilitate the training of the designed NNs, the RD spectra are stretched from 0 to 255~dB while preserving the maximum and minimum value.
In the training set, we set SCR to vary from -20 to 20~dB at 4 dB intervals, obtaining 40 sets of Monte Carlo trials for each SCR.
Consequently, we construct a training set comprising 722 RD spectra of target and 2838 RD spectra of clutter.
For the testing set, we adopt the same approach as in the training sets.
Two different testing sets are considered: the CSIR testing set, which incorporates 20 additional real CSIR sea-clutter data as backgrounds,
and IPIX testing set, comprising 1254 real IPIX sea-clutter data with distinct distributions compared to CSIR sea-clutter data.

\subsubsection{Algorithm Parameters}\label{subsubsec:Algorithm Parameters}
We assume that the process noise covariance matrix $\bm{Q}$ is known in the simulation.
The radar measurement model is
\begin{equation}\label{equ:measurement}
  \bm{z}_{j,k} = \left[ \begin{array}{ccc} 1 & 0 & 0 \\ 0 & \frac{-2f_c}{c} & 0 \\ \end{array} \right] \bm{x}_{i,k} + \bm{v}_{j,k},
\end{equation}
where $f_c$ is the radar center frequency, $c$ is the speed of light, and $\bm{v}_{j,k} \sim \mathcal{N} (\bm{0}, {\rm diag}\{(15\ {\rm m})^2,\ (0.1\ {\rm Hz})^2\}$ is a Gaussian white noise.
For DBSCAN method, the clustering thresholds of range bin and Doppler bin are set as $R_{\rm th}=45\ {\rm m}$ and $D_{\rm th}=0.3 \ {\rm Hz}$,  respectively, and the minimum number of detections required to identify a cluster is five.
For BP in the data association, we use the following parameter settings: the iterative convergence threshold is $\delta_T^{\rm BP}=10^{-6}$, the maximum number of iterations is $r_{\rm max}^{\rm BP}=1000$.
Note that $N_T$ denote the maximum numbers of targets.
A track is confirmed if $p(s_{i,k}=1)$ is greater than $\delta ^{\rm t}=0.5$ in at least three scans out of five successive scans.
A track is terminated if $p(s_{i,k}=1)$ is less than $\delta ^{\rm t}=0.5$ for three successive scans.
In case the target is visible, we empirically set the detection probability $P_{\rm d}(s_{i,k}=1)=0.9$, since most targets can be detected at a low detection threshold.
In case the target is not visible, we set the detection probability $P_{\rm d}(s_{i,k}=0)=0.01$.
The transition probability of target visibility state is set as $[\bm{T}]_{1,1} = [\bm{T}]_{2,2} = 0.85$ and $[\bm{T}]_{1,2} = [\bm{T}]_{2,1} = 0.15$.
Regarding track initialization, we impose a constraint that the number of consecutive missing measurements of any tracks should be less than $L_{\rm max}=3$.
Finally, the initial target visibility state is set to $f_{\rm s}=0.5$.

\subsubsection{Performance Evaluation}\label{subsubsec:Performance Evaluation}
The performance metrics used for evaluation are as follows.
\begin{itemize}
  \item Average multi-object tracking accuracy (AMOT)~\cite{Weng2020};
  \item Number of identity switches (IDS);
  \item Track fragments (Frag);
  \item Root mean squared error of target kinematic state (RMSE);
  \item Mean optimal subpattern assignment~\cite{OSPA200} (MOSPA) using Mahalanobis distance with the covariance matrix being ${\rm diag}\{(15\ {\rm m})^2,\ (0.1\ {\rm Hz})^2\}$, order being $p=2$ and cutoff being $c=9.4$.
\end{itemize}
The values of the performance metrics are averaged over 40 Monte Carlo runs.

\subsection{Results of MTT}\label{subsec:Results of Target Tracking}
\begin{figure}[!htbp]
  \centering
  \subfloat[\label{fig:Primary detection1-CSIR}]{\includegraphics[width=4.2cm]{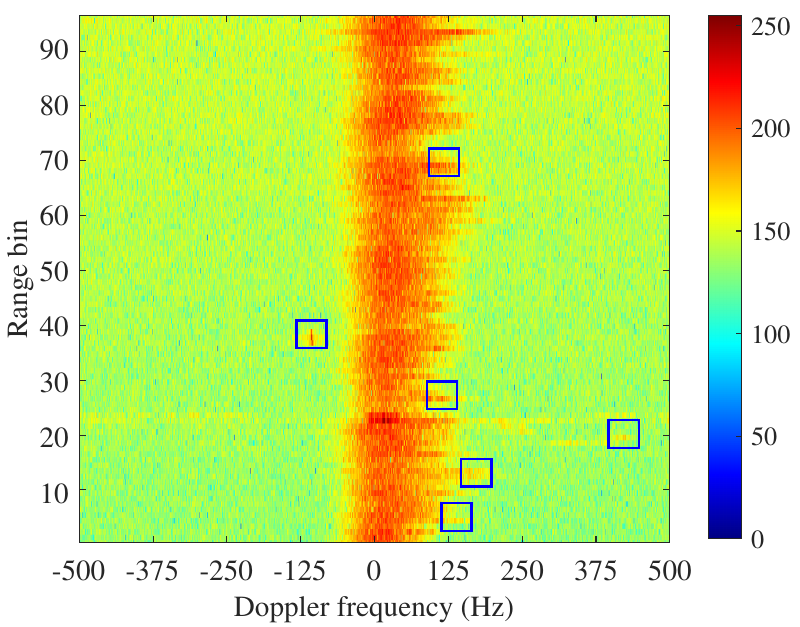}} \
  \subfloat[\label{fig:Primary detection2-CSIR}]{\includegraphics[width=4.2cm]{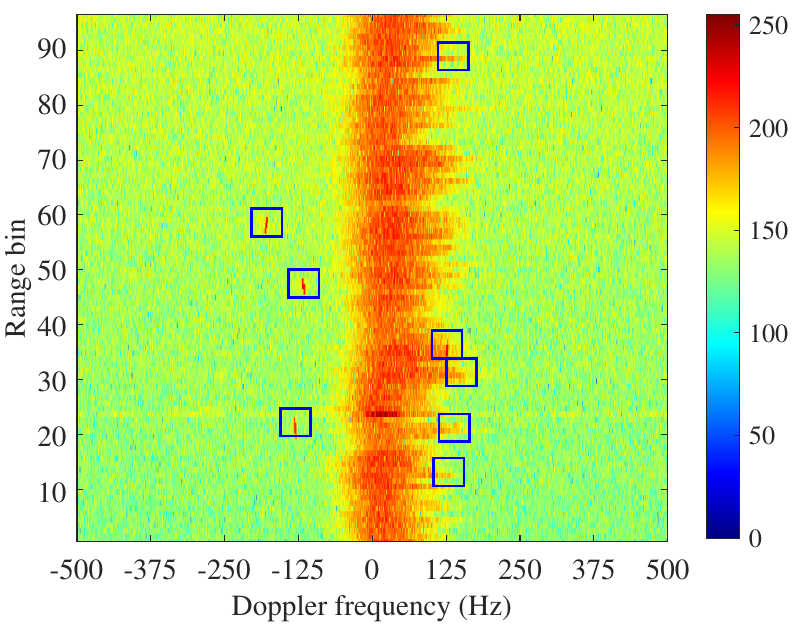}} \\
  \subfloat[\label{fig:Primary detection3-CSIR}]{\includegraphics[width=4.2cm]{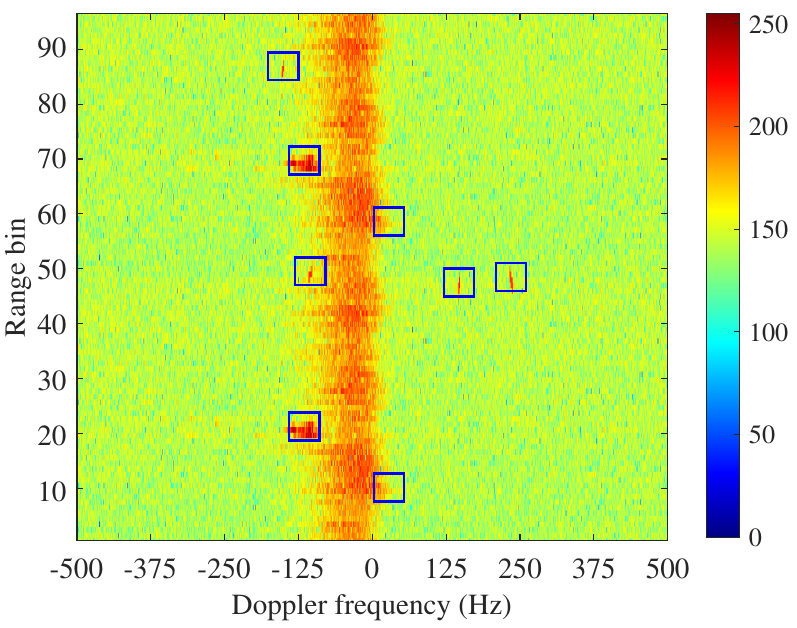}} \
  \subfloat[\label{fig:Primary detection4-CSIR}]{\includegraphics[width=4.2cm]{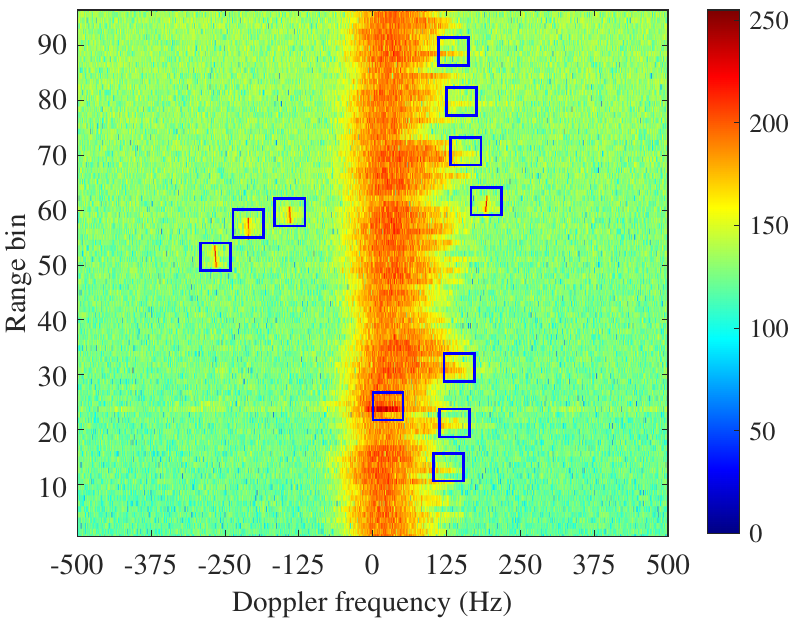}}
  \caption{Primary detection by the CFAR detector with a primary false alarm rate $P_{\rm FA}=0.28$. The targets are indicated by the blue rectangle boxes.}\label{fig:Primary detection-CSIR}
\end{figure}

The primary detection results obtained from CFAR detector and DBSCAN cluster with a primary false alarm rate $P_{\rm FA}=0.28$ are illustrated in Fig.~\ref{fig:Primary detection-CSIR}.
Additionally, Fig.~\ref{fig:Primary detection1-CSIR}-Fig.~\ref{fig:Primary detection4-CSIR} show the target detection results for different SCR of -8~dB, -4~dB, 0~dB, 4~dB, respectively.
These results demonstrate that the detector can derive a target measurement but also encounters a considerable amount of sea clutter, posing challenges for MTT.
Some targets are even located within the main sea-clutter spectra with low Doppler shift.
To improve data association performance, we utilize the RD spectra within the blue box as additional information for measurements.

\begin{figure}[!htbp]
  \centering
  \subfloat[NEMP\label{fig:Track-NEMP-CSIR}]{\includegraphics[width=7cm]{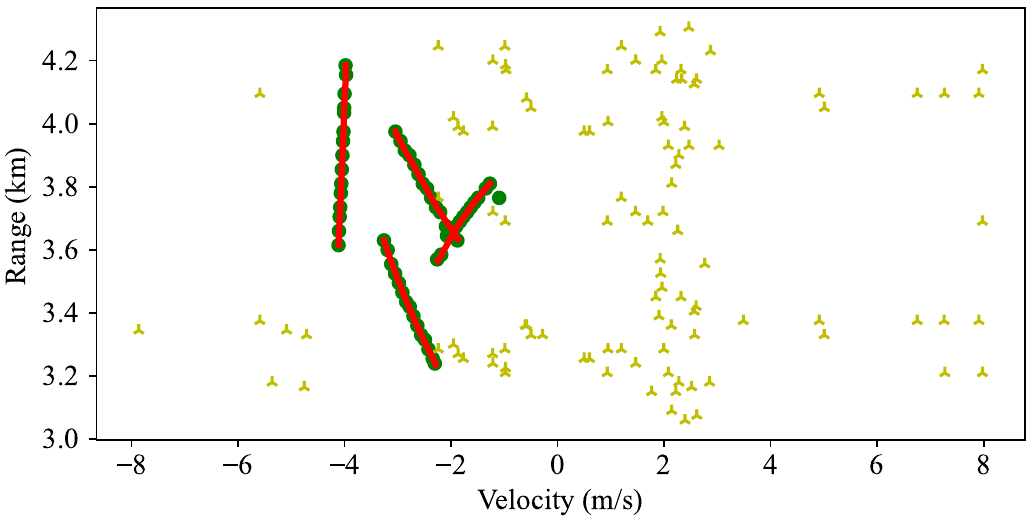}}
  \\
  \subfloat[MP\label{fig:Track-MP-CSIR}]{\includegraphics[width=8cm]{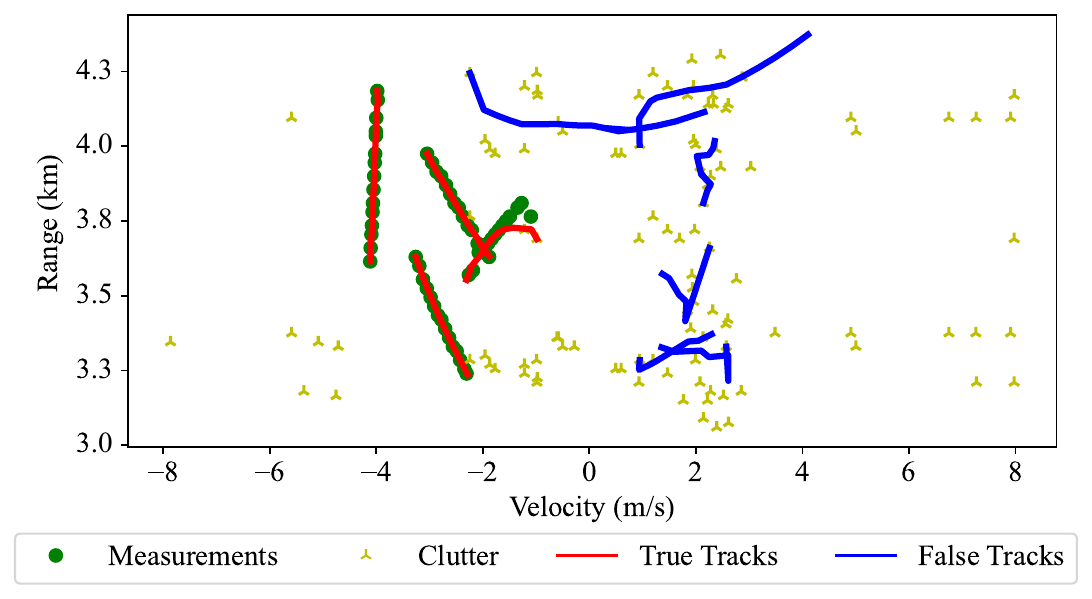}}
  \caption{Tracks obtained by NEMP and MP.}\label{fig:Estimated Trajectories of targets-CSIR}
\end{figure}

The trajectories obtained by NEMP and MP are presented in Fig.~\ref{fig:Estimated Trajectories of targets-CSIR}.
As depicted in Fig.~\ref{fig:Track-NEMP-CSIR}, NEMP successfully tracks all four targets and without generating any false tracks.
On the other hand, the tracking result of MP is shown in Fig.~\ref{fig:Track-MP-CSIR},
and it can be observed that MP can also track the four targets but generates several false tracks.
The comparison between NEMP and MP reveals that NEMP, with the aid of RD spectral feature information, significantly reduces the number of false targets in comparison to MP.

\begin{figure} [!htbp]
  \centering
  \includegraphics[width=7cm]{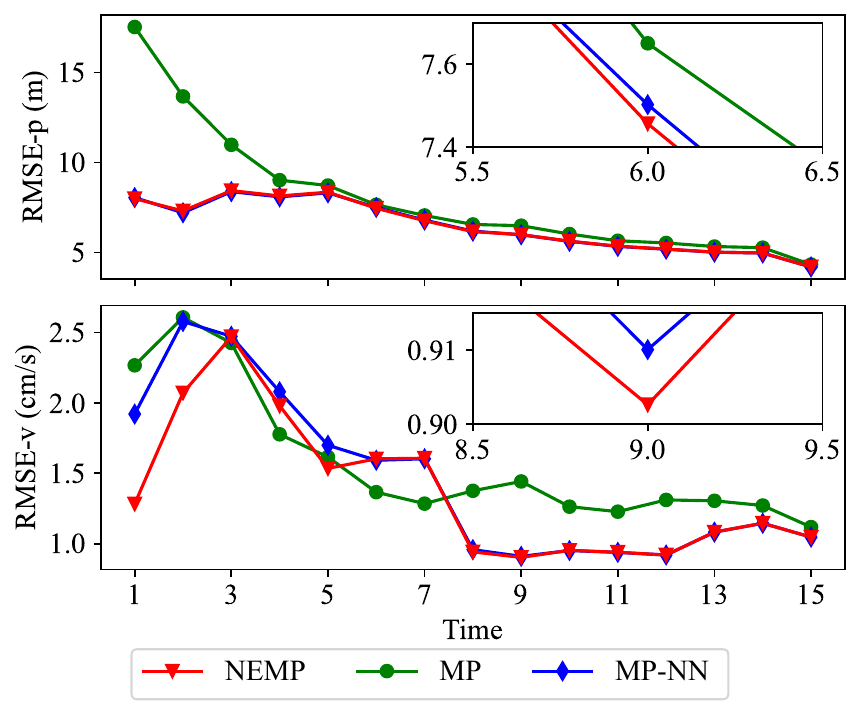} \\
  \caption{Monte Carlo average RMSE of target range and velocity versus time.}\label{fig:RMSE-CSIR}
\end{figure}

\begin{figure} [!htbp]
  \centering
  \includegraphics[width=7cm]{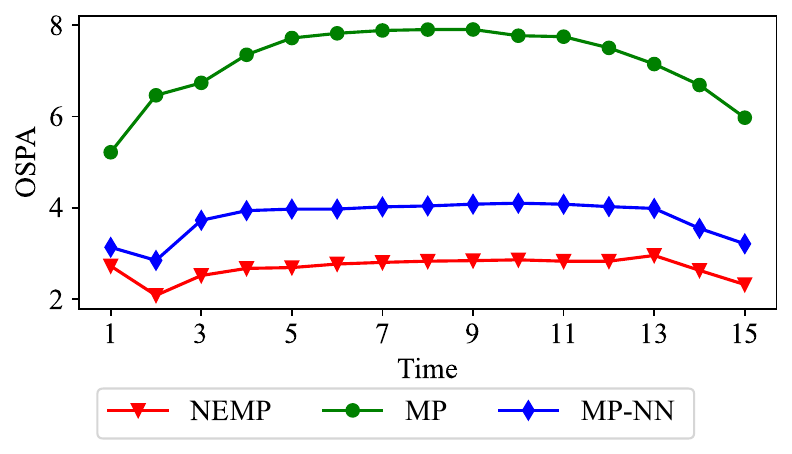} \\
  \caption{Monte Carlo average OSPA of target kinematic state versus time.}\label{fig:OSPA-CSIR}
\end{figure}

We compare the performance of NEMP with MP and MP-NN, and the corresponding results for $\rm{SCR}=\rm{0 \ dB}$ are summarized in Table~\ref{tab:Performance comparison-CSIR}.
The results indicate that NEMP achieves the best performance in terms of AMOT, followed by MP-NN, while MP shows the worst performance.
This suggests that NEMP outperforms in true track tracking, false track rejection and data association.
In terms of IDS and Frag, NEMP shows similar results to MP and MP-NN.
Fig.~\ref{fig:RMSE-CSIR} presents the Monte Carlo average RMSE of target range and velocity versus time,
with the corresponding RMSE shown in Table~\ref{tab:Performance comparison-CSIR}.
The RMSE-p metric of NEMP is slightly smaller than MP and MP-NN, while the RMSE-v metric is almost the same as MP and MP-NN.
We further report the OSPA metric versus time in Fig.~\ref{fig:OSPA-CSIR}, and MOSPA in Table~\ref{tab:Performance comparison-CSIR}.
NEMP outperforms MP and MP-NN in terms of OSPA, demonstrating its superiority in terms of localization error, false estimated tracks, and missed ground truth tracks.
\begin{table} [!htbp]
  \renewcommand \arraystretch{1.25}
  \centering
  \caption{Performance comparison.}\label{tab:Performance comparison-CSIR}
  \begin{tabular}{c|cccc}
     \hline
     \textbf{Method} & \textbf{MP} & \textbf{MP-NN} & \textbf{NEMP} \\ \hline
            {AMOT}   & 0.27         & 0.68          & \textbf{0.80}  \\
            {IDS}    & \textbf{4.00}& 4.23          & 4.18           \\
            {Frag}   & \textbf{0}   & 0.08          & 0.08           \\
    {RMSE-p (m)}    & 7.99         & 6.46          & \textbf{6.45}  \\
    {RMSE-v (cm/s)} & 1.58         & 1.46          & \textbf{1.37}  \\
            {MOSPA}  & 7.19         & 3.78          & \textbf{2.69}  \\
    \hline
  \end{tabular}
\end{table}

\begin{figure} [!htbp]
  \centering
  \includegraphics[width=8cm]{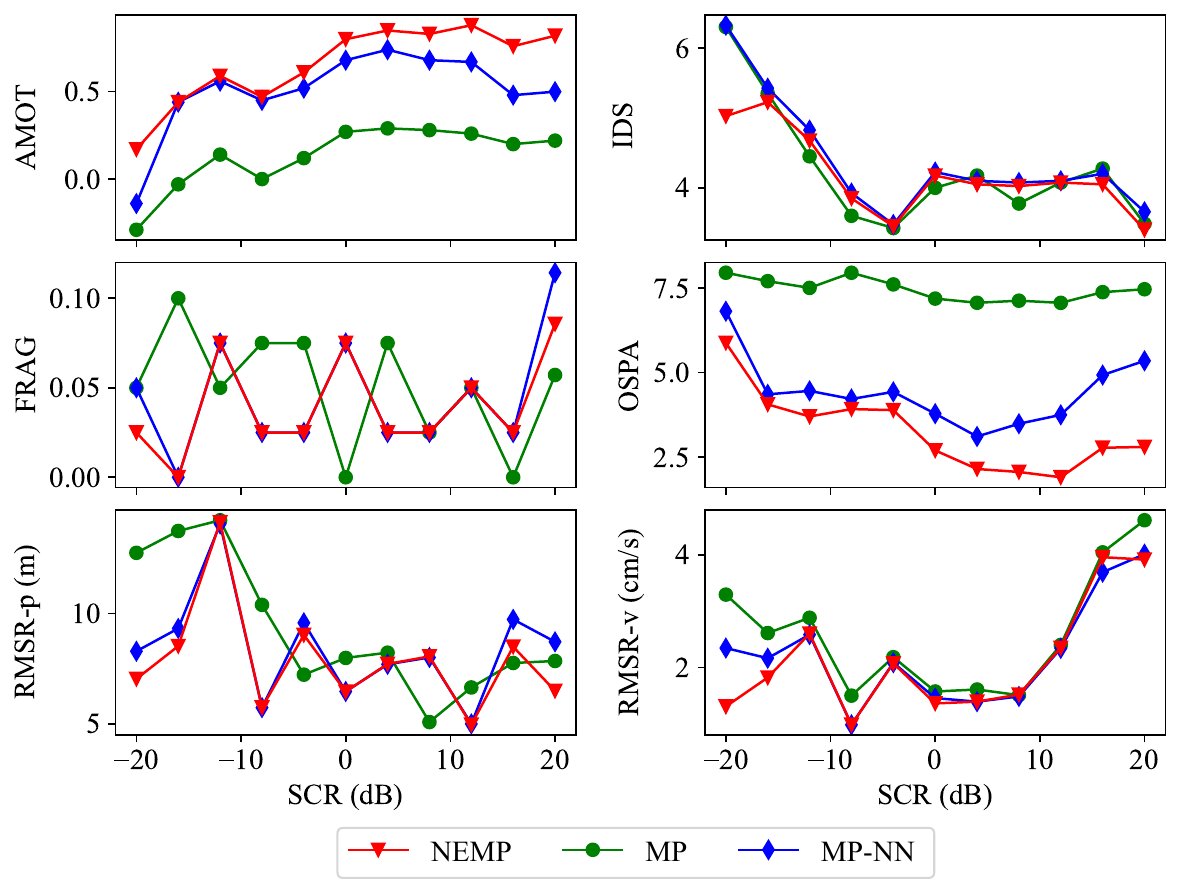} \\
  \caption{Performance comparison of MTT w.r.t different SCR.}\label{fig:MTT2SCR-CSIR}
\end{figure}

Fig.~\ref{fig:MTT2SCR-CSIR} presents the performance comparison of MTT with respect to~(w.r.t.) different SCR.
In the top right of Fig.~\ref{fig:MTT2SCR-CSIR}, it is evident that NEMP outperforms other algorithms in terms of AMOT.
This improvement is attributed to NEMP's incorporation of classification information into the MTT architecture, effectively suppressing false tracks.
Regarding IDS (as shown in the top left of Fig.~\ref{fig:MTT2SCR-CSIR}) and Frag (as shown in the middle right of Fig.~\ref{fig:MTT2SCR-CSIR}), NEMP exhibits similar results to MP and MP-NN.
In terms of OSPA (as shown in the middle left of Fig.~\ref{fig:MTT2SCR-CSIR}), NEMP performs better than MP-NN, while MP shows the worst performance.
Comparing RMSE among NEMP and other MTT methods~(as shown in the bottom of Fig.~\ref{fig:MTT2SCR-CSIR}), their performances are comparable.
Overall, NEMP outperforms the other algorithms.
This superior performance can be attributed to the use of RD-spectra features, which enhance the robustness and effectiveness of MTT.

These results clearly demonstrate the superiority of the proposed NEMP method over the MP and MP-NN methods in terms of false track rejection and overall MTT performance.
This is not surprising since NEMP incorporate additional information in the form of RD-spectra features.
In particular, the MTT method without classification aid usually assumes that false alarm measurements are uniformly distributed over the region of interest, and their occurrence is independent and identically distributed over time.
However, these assumptions often do not hold in real-world MTT applications, such as sea-surface small target tracking.
This model mismatch can lead to a degradation in tracking performance, which is effectively addressed by the false alarm rejection capability of NEMP.
The incorporation of RD-spectra features by NEMP significantly improves data association by leveraging object RD information provided by the RD features. This enhancement ensures better handling of challenging tracking scenarios and leads to more accurate and reliable tracking results.

\subsection{Generalization Ability Test}
\begin{figure}[!htbp]
  \centering
  \subfloat[\label{fig:Primary detection1-IPIX}]{\includegraphics[width=4.2cm]{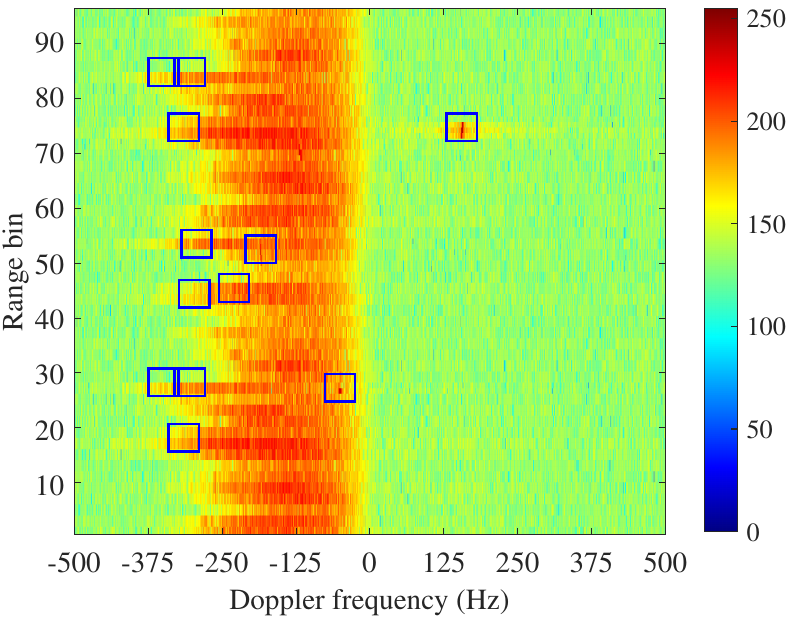}} \
  \subfloat[\label{fig:Primary detection2-IPIX}]{\includegraphics[width=4.2cm]{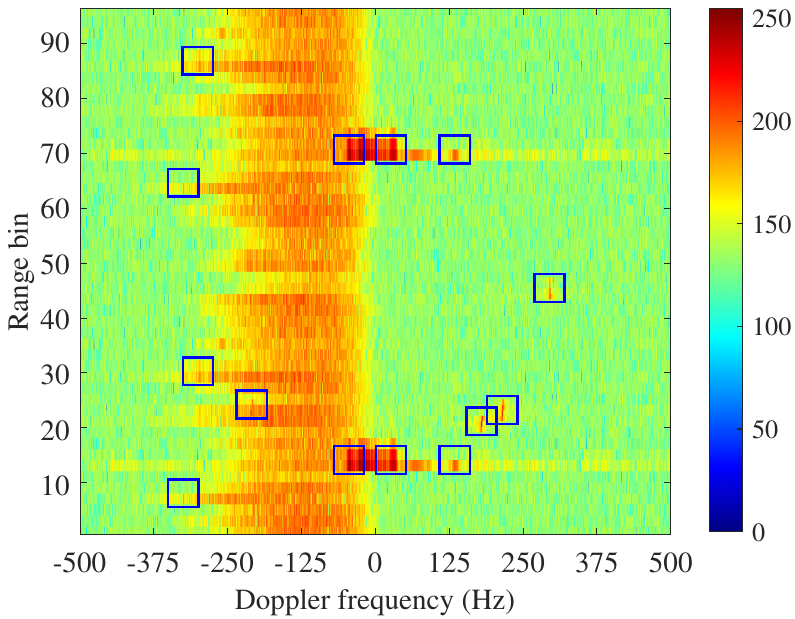}} \\
  \subfloat[\label{fig:Primary detection3-IPIX}]{\includegraphics[width=4.2cm]{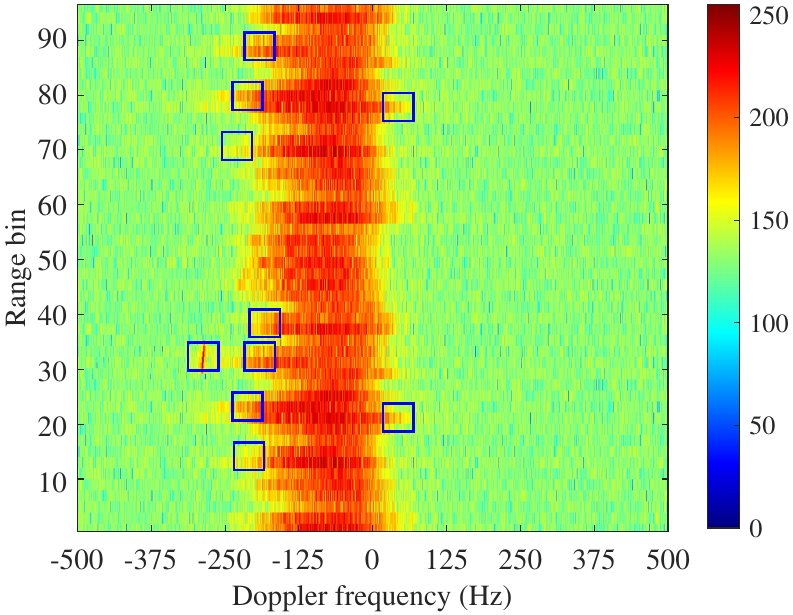}} \
  \subfloat[\label{fig:Primary detection4-IPIX}]{\includegraphics[width=4.2cm]{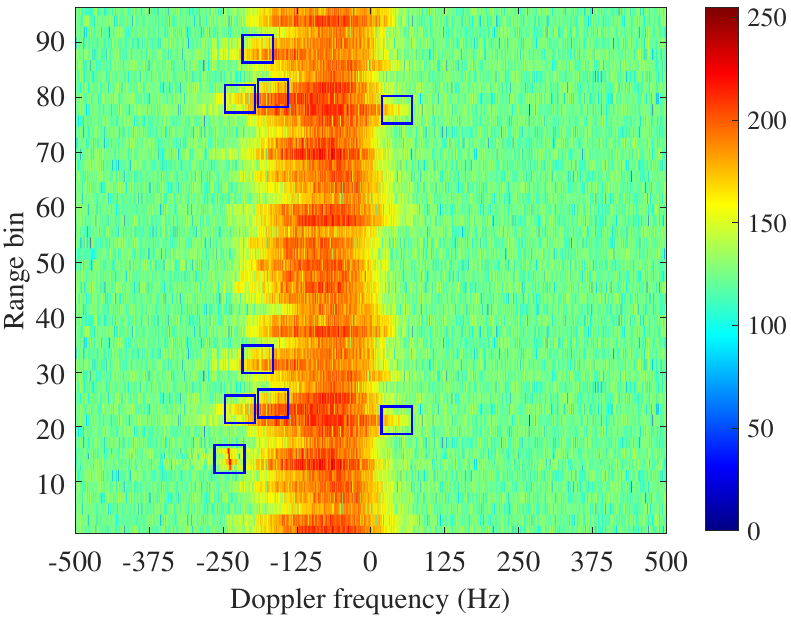}}
  \caption{Primary detection by the CFAR detector with a primary false alarm rate $P_{\rm FA}=0.28$. The targets are indicated by the blue rectangle boxes.}\label{fig:Primary detection-IPIX}
\end{figure}

We conducted validation of the proposed NEMP method using the IPIX dataset to assess its generalization capability. The IPIX dataset exhibits different Doppler characteristics and amplitude distributions of sea clutter when compared to the CSIR sea clutter used in the training set.
Fig.~\ref{fig:Primary detection1-IPIX}-Fig.~\ref{fig:Primary detection4-IPIX} illustrate the primary detection results obtained through CFAR detector and DBSCAN cluster, with a primary false alarm rate $P_{\rm FA}=0.28$, for various SCR of -8~dB, -4~dB, 0~dB, 4~dB, respectively.
As shown in Fig.~\ref{fig:Primary detection-CSIR}, a noticeable difference between the IPIX dataset and the CSIR dataset lies in the mean Doppler frequencies and spectra widths.
This discrepancy suggests distinct marine environments for the CSIR and IPIX datasets.
Notably, the IPIX dataset exhibits a much larger spectra width, resulting in a greater number of slower targets being present within the main sea clutter spectra and an increased detection of sea clutter.
This variation highlights the challenges and differences in the tracking scenarios between the two datasets and emphasizes the need for a robust and adaptable MTT method like NEMP to handle such varying marine environments effectively.

\begin{figure}[!htbp]
  \centering
  \subfloat[NEMP\label{fig:Track-NEMP-IPIX}]{\includegraphics[width=7cm]{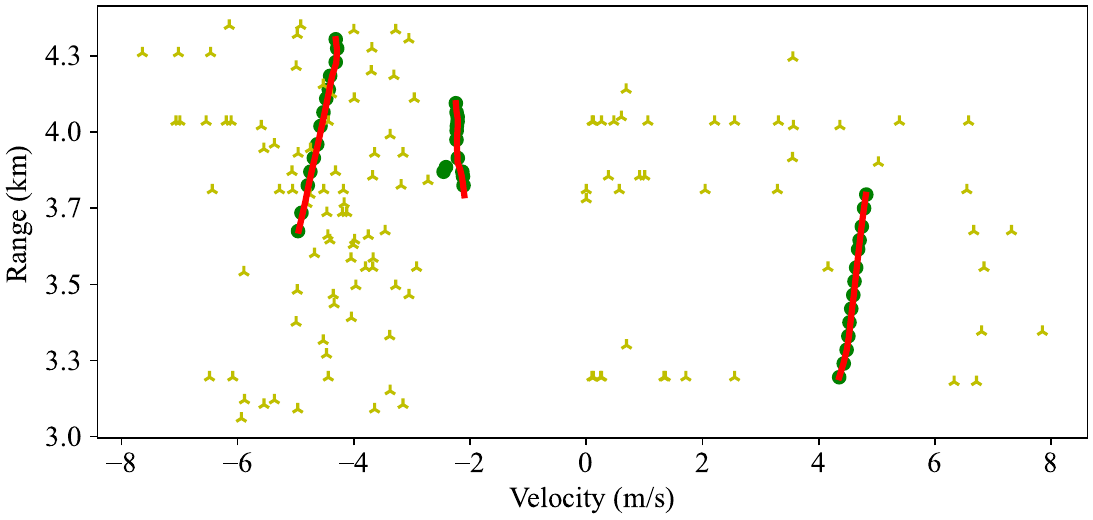}}
  \\
  \subfloat[MP\label{fig:Track-MP-IPIX}]{\includegraphics[width=7.1cm]{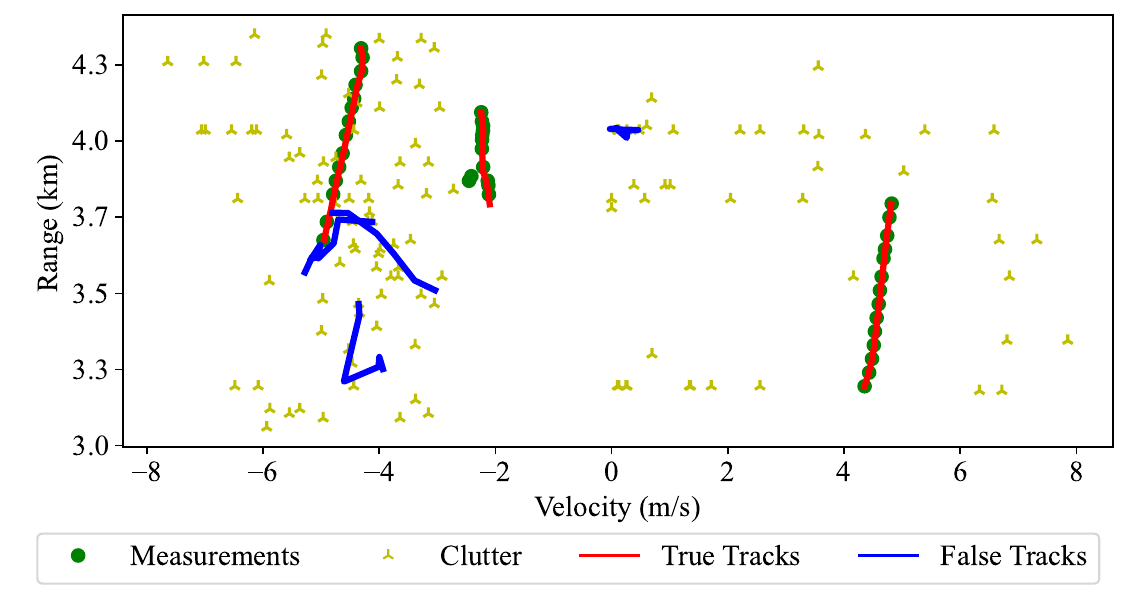}}
  \caption{Tracks obtained by NEMP and MP.}\label{fig:Estimated Trajectories of targets-IPIX}
\end{figure}

NEMP and MP are then utilized to track the targets in the IPIX dataset, and Fig.~\ref{fig:Estimated Trajectories of targets-IPIX} displays the trajectories obtained by NEMP and MP.
It can be observed that both NEMP and MP successfully track all three targets even in dense sea clutter environments.
However, MP generates several false tracks.
Similar to the test results on the CSIR dataset, NEMP effectively tracks targets in dense clutter regions and significantly reduces the number of false targets with the aid of RD spectral feature information, demonstrating its acceptable tracking performance on the IPIX testing set.

\begin{figure} [!htbp]
  \centering
  \includegraphics[width=7cm]{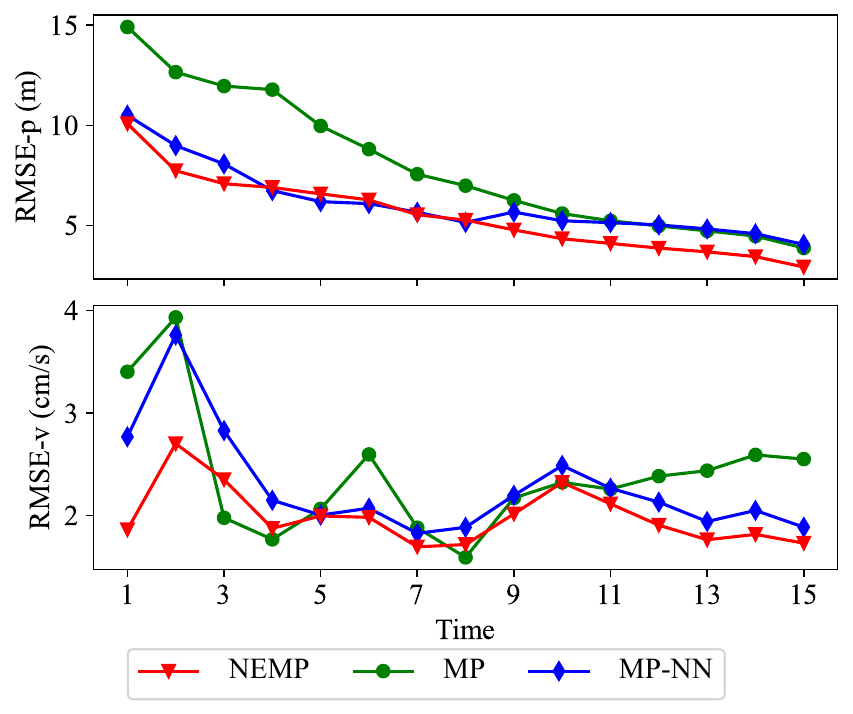} \\
  \caption{Monte Carlo average RMSE of target position versus time.}\label{fig:RMSE-IPIX}
\end{figure}

\begin{figure} [!htbp]
  \centering
  \includegraphics[width=7cm]{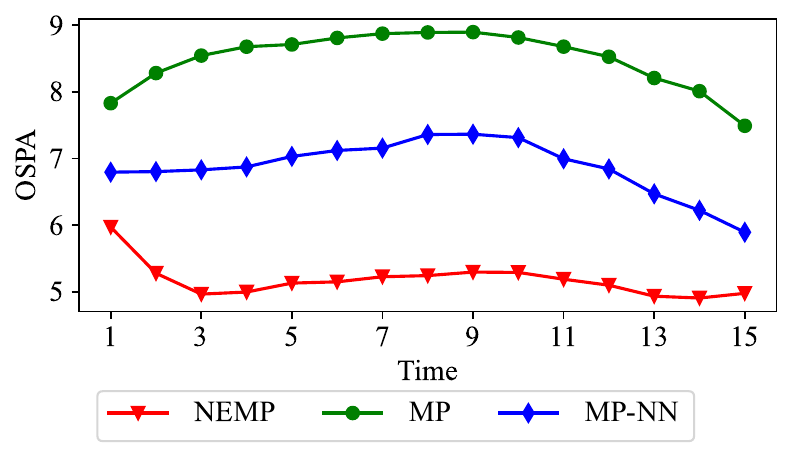} \\
  \caption{Monte Carlo average OSPA of target position versus time.}\label{fig:OSPA-IPIX}
\end{figure}

Table~\ref{tab:Performance comparison-IPIX} presents the performance metrics for MP, MP-NN, and NEMP at $\rm{SCR}=\rm{0 \ dB}$.
Consistent with the results on the CSIR dataset, NEMP outperforms MP-NN in the AMOT metric, with MP showing the worst performance.
Again, NEMP exhibits similar numbers of IDS and Frag compared to MP and MP-NN.
The Monte Carlo average RMSE and OSPA of MTT versus time are illustrated in Fig.~\ref{fig:RMSE-IPIX} and Fig.~\ref{fig:OSPA-IPIX}, respectively.
As with the CSIR dataset, NEMP slightly outperforms MP and MP-NN in terms of RMSE, while it significantly outperforms them in terms of OSPA.

Comparing Table~\ref{tab:Performance comparison-IPIX} and Table~\ref{tab:Performance comparison-CSIR}, it can be observed that the AMOT and OSPA of the proposed approach on the IPIX testing set are slightly worse than those on the CSIR testing set. This difference is attributed to the wider main sea-clutter spectra in the IPIX sea clutter data, resulting in more low-velocity targets being within the main sea-clutter spectra and going undetected.
\begin{table} [!htbp]
  \renewcommand \arraystretch{1.25}
  \centering
  \caption{Performance comparison.}\label{tab:Performance comparison-IPIX}
  \begin{tabular}{c|cccc}
     \hline
     \textbf{Method} & \textbf{MP}  & \textbf{MP-NN} & \textbf{NEMP} \\ \hline
            {AMOT}   & -0.21        & 0.06          & \textbf{0.27}  \\
            {IDS}    & \textbf{3.93}& 4.30          & 4.10           \\
            {Frag}   & 0.03         & \textbf{0}    & \textbf{0}     \\
    {RMSE-p (m)}    & 6.13         & 7.99          & \textbf{5.51}  \\
    {RMSE-v (cm/s)} & 2.40         & 2.28          & \textbf{1.99}  \\
            {MOSPA}  & 8.48         & 6.87          & \textbf{5.18}  \\
    \hline
  \end{tabular}
\end{table}

\begin{figure} [!htbp]
  \centering
  \includegraphics[width=8cm]{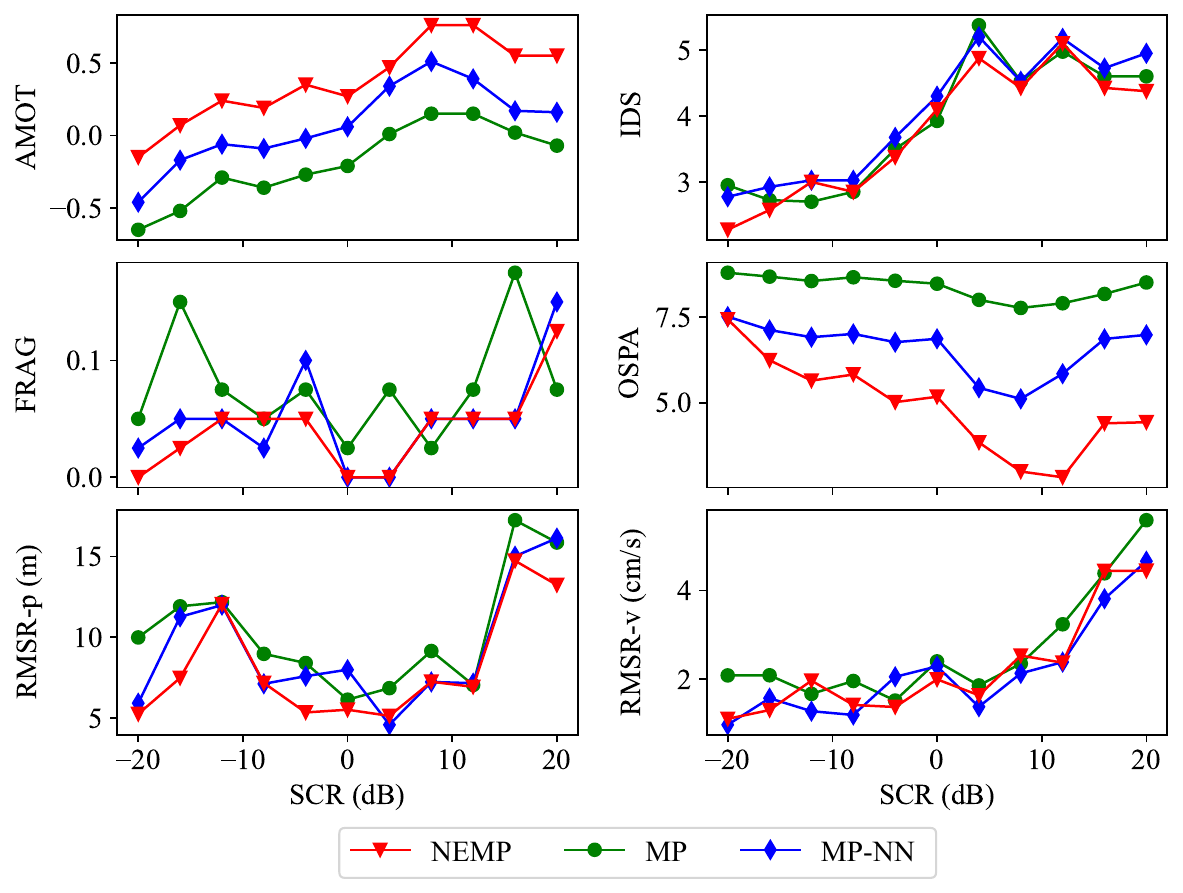} \\
  \caption{Performance comparison of MTT w.r.t different SCR.}\label{fig:MTT2SCR-IPIX}
\end{figure}

In Fig.~\ref{fig:MTT2SCR-IPIX}, we present the performance comparison of MTT w.r.t different SCR on the IPIX dataset, and the conclusions obtained are comparable to those of the CSIR dataset.
NEMP performs better than other algorithms in terms of AMOT (as shown in the top right of Fig.~\ref{fig:MTT2SCR-IPIX}). Fig.~\ref{fig:MTT2SCR-IPIX} also shows that the proposed NEMP method exhibits similar numbers of IDS and Frag compared to MP and MP-NN.
As indicated in the middle left of Fig.~\ref{fig:MTT2SCR-IPIX}, NEMP outperforms MP-NN, while MP shows the worst performance in terms of OSPA.
The bottom of Fig.~\ref{fig:MTT2SCR-IPIX} demonstrates that NEMP and other MTT methods have comparable performance on the RMSE.
It is noteworthy that, compared to the CSIR dataset, NEMP exhibits significant improvement over MP-NN on the IPIX dataset in terms of AMOT and OSPA. 
This improvement can be attributed to the complex sea surface environment of the IPIX dataset, which causes the performance of the RD spectral classifier to degrade, leading to poor MP-NN results.
NEMP integrates RD spectral classifier information into the data association, improving the robustness of the tracker through multi-frame decision making for track management.
Overall, the results demonstrate that our proposed NEMP approach achieves good tracking performance of sea-surface targets under different marine environments.

\section{Conclusions}\label{sec:CONCLUSIONS}
This article presented a classification-aided robust MTT using NEMP.
The proposed approach utilizes a unified tracking framework, incorporating three key modules: an MP module for modeling the target's kinematic state and spatial information of measurements, a NN module for modeling RD spectra information of measurements and classifying target-generated and clutter-generated measurements, and a DS module for fusing beliefs from MP and NN.
Compared to the MTT algorithm based on MP without classification aid and the MTT algorithm based on MP using measurement suppression by the RD-spectra classifier, this approach has demonstrated superior performance in tracking small targets with reduced false tracks.
the incorporation of classification aid significantly enhanced the tracking accuracy.
Moreover, the proposed approach exhibited good generalization ability, which is crucial for deep learning-based algorithms developed for radar applications.


\bibliographystyle{IEEEtran}
\bibliography{IEEEabrv,FAD-MTT}

\begin{thebibliography}{10}
\providecommand{\url}[1]{#1}
\csname url@samestyle\endcsname
\providecommand{\newblock}{\relax}
\providecommand{\bibinfo}[2]{#2}
\providecommand{\BIBentrySTDinterwordspacing}{\spaceskip=0pt\relax}
\providecommand{\BIBentryALTinterwordstretchfactor}{4}
\providecommand{\BIBentryALTinterwordspacing}{\spaceskip=\fontdimen2\font plus
\BIBentryALTinterwordstretchfactor\fontdimen3\font minus
  \fontdimen4\font\relax}
\providecommand{\BIBforeignlanguage}[2]{{%
\expandafter\ifx\csname l@#1\endcsname\relax
\typeout{** WARNING: IEEEtran.bst: No hyphenation pattern has been}%
\typeout{** loaded for the language `#1'. Using the pattern for}%
\typeout{** the default language instead.}%
\else
\language=\csname l@#1\endcsname
\fi
#2}}
\providecommand{\BIBdecl}{\relax}
\BIBdecl

\bibitem{2011Tracking}
Y.~Bar-Shalom, P.~Willett, and X.~Tian, \emph{Tracking and Data Fusion: A
  Handbook of Algorithms}.\hskip 1em plus 0.5em minus 0.4em\relax Bloomfield,
  CT : YBS Publishing, 2011.

\bibitem{Blackman2004}
S.~S. Blackman, ``{Multiple hypothesis tracking for multiple target
  tracking},'' \emph{IEEE Aerospace and Electronic Systems Magazine}, vol.~19,
  no.~1, pp. 5--18, 2004.

\bibitem{Bar1995Multitarget}
Y.~Bar-Shalom and X.~Li, \emph{Multitarget-multisensor tracking: principles and
  techniques}.\hskip 1em plus 0.5em minus 0.4em\relax Storrs CT : YBS
  publishing, 1995.

\bibitem{Vo2007}
B.-T. Vo, B.-N. Vo, and A.~Cantoni, ``{Analytic implementations of the
  cardinalized probability hypothesis density filter},'' \emph{IEEE
  Transactions on Signal Processing}, vol.~55, no.~13, pp. 3553--3567, 2007.

\bibitem{Vo2009}
------, ``{The cardinality balanced multitarget multi-Bernoulli filter and its
  implementations},'' \emph{IEEE Transactions on Signal Processing}, vol.~57,
  no.~2, pp. 409--423, 2009.

\bibitem{Grossi2013}
E.~Grossi, M.~Lops, and L.~Venturino, ``{A novel dynamic programming algorithm
  for track-before-detect in radar systems},'' \emph{IEEE Transactions on
  Signal Processing}, vol.~61, no.~10, pp. 2608--2619, 2013.

\bibitem{Aprile2016}
A.~Aprile, E.~Grossi, M.~Lops, and L.~Venturino, ``{Track-before-detect for sea
  clutter rejection: Tests with real data},'' \emph{IEEE Transactions on
  Aerospace and Electronic Systems}, vol.~52, no.~3, pp. 1035--1045, 2016.

\bibitem{Richards2010}
M.~A. Richards, J.~Scheer, W.~A. Holm, and W.~L. Melvin, \emph{Principles of
  modern radar: basic principles}.\hskip 1em plus 0.5em minus 0.4em\relax NC
  Raleigh: SciTech Publising, 2010.

\bibitem{Li2014}
Y.~Li, G.~Zhang, and R.~J. Doviak, ``Ground clutter detection using the
  statistical properties of signals received with a polarimetric radar,''
  \emph{IEEE Transactions on Signal Processing}, vol.~62, no.~3, pp. 597--606,
  2014.

\bibitem{Shi2019}
S.~N. Shi, X.~Liang, P.~L. Shui, J.~K. Zhang, and S.~Zhang, ``{Low-velocity
  small target detection with doppler-guided retrospective filter in
  high-resolution aadar at fast scan mode},'' \emph{IEEE Transactions on
  Geoscience and Remote Sensing}, vol.~57, no.~11, pp. 8937--8953, 2019.

\bibitem{Shui2020}
P.~Shui, Z.~Guo, and S.~Shi, ``{Feature-compression-based detection of
  sea-surface small targets},'' \emph{IEEE Access}, vol.~8, pp. 8371--8385,
  2020.

\bibitem{Gao2021}
C.~Gao, J.~Yan, X.~Peng, and H.~Liu, ``{Signal structure information-based
  target detection with a fully convolutional network},'' \emph{Information
  Sciences}, vol. 576, pp. 345--354, 2021.

\bibitem{Shalom2005}
Y.~Bar-Shalom, T.~Kirubarajan, and C.~Gokberk, ``Tracking with
  classification-aided multiframe data association,'' \emph{IEEE Transactions
  on Aerospace and Electronic Systems}, vol.~41, no.~3, pp. 868--878, 2005.

\bibitem{Wen2022}
L.~Wen, J.~Ding, and Z.~Xu, ``{Multiframe detection of sea-surface small target
  using deep convolutional neural network},'' \emph{IEEE Transactions on
  Geoscience and Remote Sensing}, vol.~60, pp. 1--16, 2022.

\bibitem{Yedidia2005}
J.~S. Yedidia, W.~T. Freeman, and Y.~Weiss, ``{Constructing free-energy
  approximations and generalized belief propagation algorithms},'' \emph{IEEE
  Transactions on Information Theory}, vol.~51, no.~7, pp. 2282--2312, 2005.

\bibitem{Zhang2019}
C.~Zhang, J.~Butepage, H.~Kjellstrom, and S.~Mandt, ``{Advances in variational
  inference},'' \emph{IEEE Transactions on Pattern Analysis and Machine
  Intelligence}, vol.~41, no.~8, pp. 2008--2026, 2019.

\bibitem{Riegler2013}
E.~Riegler, G.~E. Kirkelund, C.~N. Manch{\'{o}}n, M.~A. Badiu, and B.~H.
  Fleury, ``{Merging belief propagation and the mean field approximation: A
  free energy approach},'' \emph{IEEE Transactions on Information Theory},
  vol.~59, no.~1, pp. 588--602, 2013.

\bibitem{Williams2014}
J.~L. Williams and R.~A. Lau, ``{Approximate evaluation of marginal association
  probabilities with belief propagation},'' \emph{IEEE Transactions on
  Aerospace and Electronic Systems}, vol.~50, no.~4, pp. 2942--2959, 2014.

\bibitem{Williams2018}
------, ``{Multiple scan data association by convex variational inference},''
  \emph{IEEE Transactions on Signal Processing}, vol.~66, no.~8, pp.
  2112--2127, 2018.

\bibitem{Sun2016IF}
S.~Sun, H.~Lan, Z.~Wang, Q.~Pan, and H.~Zhang, ``{The application of
  sum-product algorithm for data association},'' in \emph{Proceedings of 19th
  International Conference on Information Fusion}.\hskip 1em plus 0.5em minus
  0.4em\relax ISIF, 2016, pp. 416--423.

\bibitem{meyer2017scalable}
F.~Meyer, P.~Braca, P.~Willett, and F.~Hlawatsch, ``{A scalable algorithm for
  tracking an unknown number of targets using multiple sensors},'' \emph{IEEE
  Transactions on Signal Processing}, vol.~65, no.~13, pp. 3478--3493, 2017.

\bibitem{Meyer2018}
F.~Meyer, T.~Kropfreiter, J.~L. Williams, R.~Lau, F.~Hlawatsch, P.~Braca, and
  M.~Z. Win, ``{Message passing algorithms for scalable multitarget
  tracking},'' \emph{Proceedings of the IEEE}, vol. 106, no.~2, pp. 121--259,
  2018.

\bibitem{Soldi2019}
G.~Soldi, F.~Meyer, P.~Braca, and F.~Hlawatsch, ``{Self-tuning algorithms for
  multisensor-multitarget tracking using belief propagation},'' \emph{IEEE
  Transactions on Signal Processing}, vol.~67, no.~15, pp. 3922--3937, 2019.

\bibitem{Kropfreiter2019}
T.~Kropfreiter, F.~Meyer, and F.~Hlawatsch, ``A fast labeled {multi-Bernoulli}
  filter using belief propagation,'' \emph{IEEE Transactions on Aerospace and
  Electronic Systems}, vol.~56, no.~3, pp. 2478--2488, 2019.

\bibitem{Sharma2019}
P.~Sharma, A.~A. Saucan, D.~J. Bucci, and P.~K. Varshney, ``Decentralized
  gaussian filters for cooperative self-localization and multi-target
  tracking,'' \emph{IEEE Transactions on Signal Processing}, vol.~67, no.~22,
  pp. 5896--5911, 2019.

\bibitem{Meyer2020}
F.~Meyer and M.~Z. Win, ``{Scalable data association for extended object
  tracking},'' \emph{IEEE Transactions on Signal and Information Processing
  over Networks}, vol.~6, pp. 491--507, 2020.

\bibitem{meyer2021scalable}
F.~Meyer and J.~L. Williams, ``Scalable detection and tracking of geometric
  extended objects,'' \emph{IEEE Transactions on Signal Processing}, vol.~69,
  pp. 6283--6298, 2021.

\bibitem{Leitinger2019a}
E.~Leitinger, F.~Meyer, F.~Hlawatsch, K.~Witrisal, F.~Tufvesson, and M.~Z. Win,
  ``{A belief propagation algorithm for multipath-based SLAM},'' \emph{IEEE
  Transactions on Wireless Communications}, vol.~18, no.~12, pp. 5613--5629,
  2019.

\bibitem{Li2022}
X.~Li, E.~Leitinger, A.~Venus, and F.~Tufvesson, ``{Sequential detection and
  estimation of multipath channel parameters using belief propagation},''
  \emph{IEEE Transactions on Wireless Communications}, vol.~21, no.~10, pp.
  8385--8402, 2022.

\bibitem{Cormack2019}
D.~Cormack, I.~Schlangen, J.~R. Hopgood, and D.~E. Clark, ``{Joint registration
  and fusion of an infra-red camera and scanning radar in a maritime
  context},'' \emph{IEEE Transactions on Aerospace and Electronic Systems},
  vol.~56, no.~2, pp. 1357--1369, 2019.

\bibitem{Gaglione2022}
D.~Gaglione, P.~Braca, G.~Soldi, F.~Meyer, F.~Hlawatsch, and M.~Z. Win,
  ``{Fusion of sensor measurements and target-provided information in
  multitarget tracking},'' \emph{IEEE Transactions on Signal Processing},
  vol.~70, pp. 322--336, 2022.

\bibitem{Turner2014}
R.~D. Turner, S.~Bottone, and B.~Avasarala, ``A complete variational tracker,''
  in \emph{Advances in Neural Information Processing Systems}, vol.~27.\hskip
  1em plus 0.5em minus 0.4em\relax Curran Associates, Inc., 2014, pp. 496--504.

\bibitem{Lau2016}
R.~A. Lau and J.~L. Williams, ``{A structured mean field approach for
  existence-based multiple target tracking},'' in \emph{Proceedings of 19th
  International Conference on Information Fusion}.\hskip 1em plus 0.5em minus
  0.4em\relax ISIF, 2016, pp. 1111--1118.

\bibitem{Lan2019}
H.~Lan, S.~Sun, Z.~Wang, Q.~Pan, and Z.~Zhang, ``{Joint target detection and
  tracking in multipath environment: A variational Bayesian approach},''
  \emph{IEEE Transactions on Aerospace and Electronic Systems}, vol.~56, no.~3,
  pp. 2136--2156, 2020.

\bibitem{Lan-2020-107621}
H.~Lan, J.~Ma, Z.~Wang, Q.~Pan, and X.~Xu, ``{A message passing approach for
  multiple maneuvering target tracking},'' \emph{Signal Processing}, vol. 174,
  p. 107621, 2020.

\bibitem{Lan2020}
H.~Lan, Z.~Wang, X.~Bai, Q.~Pan, and K.~Lu, ``{Measurement-level target
  tracking fusion for over-the-horizon radar network using message passing},''
  \emph{IEEE Transactions on Aerospace and Electronic Systems}, vol.~57, no.~3,
  pp. 1600--1623, 2021.

\bibitem{Johnson2016}
M.~J. Johnson, D.~Duvenaud, A.~B. Wiltschko, S.~R. Datta, and R.~P. Adams,
  ``{Composing graphical models with neural networks for structured
  representations and fast inference},'' in \emph{Advances in Neural
  Information Processing Systems}, vol.~29.\hskip 1em plus 0.5em minus
  0.4em\relax Curran Associates, Inc., 2016, pp. 2954--2962.

\bibitem{Kuck2020}
J.~Kuck, S.~Chakraborty, H.~Tang, R.~Luo, J.~Song, A.~Sabharwal, and S.~Ermon,
  ``{Belief propagation neural networks},'' in \emph{Advances in Neural
  Information Processing Systems}, vol.~33.\hskip 1em plus 0.5em minus
  0.4em\relax Curran Associates, Inc., 2020, pp. 667--678.

\bibitem{Satorras2019}
V.~G. Satorras, Z.~Akata, and M.~Welling, ``{Combining generative and
  discriminative models for hybrid inference},'' in \emph{Advances in Neural
  Information Processing Systems}, vol.~32.\hskip 1em plus 0.5em minus
  0.4em\relax Curran Associates, Inc., 2019, p. 13802–13812.

\bibitem{Satorras2020}
V.~G. Satorras and M.~Welling, ``{Neural enhanced belief propagation on factor
  graphs},'' in \emph{International Conference on Artificial Intelligence and
  Statistics}, vol. 130.\hskip 1em plus 0.5em minus 0.4em\relax Curran
  Associates, Inc., 2020, pp. 685--693.

\bibitem{Liang2021}
M.~Liang and F.~Meyer, ``Neural enhanced belief propagation for cooperative
  localization,'' in \emph{2021 IEEE Statistical Signal Processing Workshop
  (SSP)}.\hskip 1em plus 0.5em minus 0.4em\relax IEEE, 2021, pp. 326--330.

\bibitem{Liang2022Fusion}
------, ``{Neural enhanced belief propagation for data association in
  multiobject tracking},'' in \emph{Proceedings of 25th International
  Conference on Information Fusion}.\hskip 1em plus 0.5em minus 0.4em\relax
  ISIF, 2022, pp. 1--7.

\bibitem{Liang2022}
------, ``Neural enhanced belief propagation for multiobject tracking,''
  \emph{arXiv preprint arXiv:2212.08340}, 2022.

\bibitem{Soldi2020}
G.~Soldi, D.~Gaglione, G.~{De Magistris}, P.~Braca, P.~Stinco, G.~Ferri,
  A.~Tesei, and K.~{Le Page}, ``Underwater tracking based on the sum-product
  algorithm enhanced by a neural network detections classifier,''
  \emph{Proceedings of IEEE International Conference on Acoustics, Speech and
  Signal Processing}, pp. 5460--5464, 2020.

\bibitem{Gaglione2020}
D.~Gaglione, G.~Soldi, P.~Braca, G.~{De Magistris}, F.~Meyer, and F.~Hlawatsch,
  ``Classification-aided multitarget tracking using the sum-product
  algorithm,'' \emph{IEEE Signal Processing Letters}, vol.~27, pp. 1710--1714,
  2020.

\bibitem{Liu-Zhun-Ga2018}
Z.~Liu, Q.~Pan, J.~Dezert, and A.~Martin, ``Combination of classifiers with
  optimal weight based on evidential reasoning,'' \emph{IEEE Transactions on
  Fuzzy Systems}, vol.~26, no.~3, pp. 1217--1230, 2018.

\bibitem{Huang2019}
P.~Huang, X.-G. Xia, G.~Liao, Z.~Yang, and Y.~Zhang, ``{Long-time coherent
  integration algorithm for radar maneuvering weak target with acceleration
  rate},'' \emph{IEEE Transactions on Geoscience and Remote Sensing}, vol.~57,
  no.~6, pp. 3528--3542, 2019.

\bibitem{Li2013}
Y.~Li, G.~Zhang, R.~J. Doviak, L.~Lei, and Q.~Cao, ``A new approach to detect
  ground clutter mixed with weather signals,'' \emph{IEEE Transactions on
  Geoscience and Remote Sensing}, vol.~51, no.~4, pp. 2373--2387, 2013.

\bibitem{Lecun2015}
Y.~Lecun, Y.~Bengio, and G.~Hinton, ``{Deep learning},'' \emph{Nature}, vol.
  521, no. 7553, pp. 436--444, 2015.

\bibitem{Oh2009}
S.~Oh, S.~Russell, and S.~Sastry, ``{Markov chain Monte Carlo data association
  for multi-target tracking},'' \emph{IEEE Transactions on Automatic Control},
  vol.~54, no.~3, pp. 481--497, 2009.

\bibitem{Wind2010}
H.~J.~D. Wind, J.~E. Cilliers, and P.~L. Herselman, ``{Dataware: sea clutter
  and small boat radar reflectivity databases},'' \emph{IEEE Signal Processing
  Magazine}, vol.~27, no.~2, pp. 145--148, 2010.

\bibitem{Bakker2019}
\BIBentryALTinterwordspacing
R.~Bakker and B.~Currie. {The McMaster IPIX radar sea clutter database}. Jul.
  2023. [Online]. Available:
  \url{{http://soma.ece.mcmaster.ca/ipix/grimsby/index.html}}
\BIBentrySTDinterwordspacing

\bibitem{DBSCAN}
M.~Ester, H.-P. Kriegel, J.~Sander, and X.~Xu, ``A density-based algorithm for
  discovering clusters in large spatial databases with noise,'' in
  \emph{Proceedings of the Second International Conference on Knowledge
  Discovery and Data Mining}, ser. KDD'96.\hskip 1em plus 0.5em minus
  0.4em\relax AAAI Press, 1996, pp. 226--231.

\bibitem{Weng2020}
X.~Weng, J.~Wang, D.~Held, and K.~Kitani, ``{3D multi-object tracking: A
  baseline and new evaluation metrics},'' in \emph{2020 IEEE/RSJ International
  Conference on Intelligent Robots and Systems (IROS)}, 2020, pp.
  10\,359--10\,366.

\bibitem{OSPA200}
D.~Schuhmacher, B.~T. Vo, and B.~N. Vo, ``{A consistent metric for performance
  evaluation of multi-object filters},'' \emph{IEEE Transactions on Signal
  Processing}, vol.~56, no.~8, pp. 3447--3457, 2008.

\end{thebibliography}

\end{document}